\documentclass[final,1p,times]{elsarticle}
\usepackage{amssymb}
\usepackage{amsmath}

\usepackage{xurl}
\usepackage{hyperref}
\hypersetup{colorlinks, allcolors=black}
\usepackage{epsfig}
\usepackage{comment} 
\usepackage{algorithm2e}
\usepackage{subcaption}
\usepackage{amsthm}
\usepackage{multirow}
\newtheorem{definition}{Definition}[section]
\newtheorem{lemma}{Lemma}

\usepackage{nicefrac}
\usepackage{booktabs}
\usepackage{graphicx} 
\usepackage{comment}

\RestyleAlgo{ruled}
\SetKwComment{Comment}{/* }{ */}
\SetKwInput{KwData}{Input}
\SetKwInput{KwResult}{Output}
\journal{Electronic Commerce Research and Applications}

\begin{document}

\begin{frontmatter}

\title{DP-SGD-Global-Adapt-V2-S: Triad Improvements of Privacy, Accuracy and Fairness via Step Decay Noise Multiplier  and Step Decay Upper Clipping Threshold}

\author[label1]{Sai Venkatesh Chilukoti}
\author[label1]{Md Imran Hossen}
\author[label1]{Liqun Shan}
\author[label2]{Vijay Srinivas Tida}
\author[label3]{Mahathir Mohammad Bappy} 
\author[label3]{ Wenmeng Tian}
\author[label1]{Xiali Hei}

\affiliation[label1]{organization={University of Louisiana at Lafayette},
city={Lafayette},
state={LA},
country={United States }}

\affiliation[label2]{organization={College of Saint Benedict and Saint John’s University},
city={ St. Joseph},
state={MN},
country={United States}}

\affiliation[label3]{organization={Mississippi State University},
city={ Starkville},
state={MS},
country={United States}}

%% Abstract
\begin{abstract}
Differentially Private Stochastic Gradient Descent (DP-SGD) has become a widely used technique for safeguarding sensitive information in deep learning applications. Unfortunately, DP-SGD's per-sample gradient clipping and uniform noise addition during training can significantly degrade model utility and fairness. We observe that the latest DP-SGD-Global-Adapt average gradient norm is the same throughout the training. Even when it is integrated with the existing linear decay noise multiplier, it has little or no advantage. Moreover, we notice that its upper clipping threshold increases exponentially towards the end of training, potentially impacting the model's convergence. Other algorithms, DP-PSAC, Auto-S, DP-SGD-Global, and DP-F, have utility and fairness that are similar to or worse than DP-SGD, as demonstrated in experiments. To overcome these problems and improve utility and fairness, we developed the DP-SGD-Global-Adapt-V2-S. It has a step-decay noise multiplier and an upper clipping threshold that is also decayed step-wise. DP-SGD-Global-Adapt-V2-S with a privacy budget ($\epsilon$) of 1 improves accuracy by 0.9795\%, 0.6786\%, and 4.0130\% in MNIST, CIFAR10, and CIFAR100, respectively. It also reduces the privacy cost gap ($\pi$) by 89.8332\% and 60.5541\% in unbalanced MNIST and Thinwall datasets, respectively. Finally, we develop mathematical expressions to compute the privacy budget using truncated concentrated differential privacy (tCDP) for DP-SGD-Global-Adapt-V2-T and DP-SGD-Global-Adapt-V2-S. 

\end{abstract}


%item Research highlight 1

%% Keywords
\begin{keyword}
DP-SGD-Global-Adapt-V2 \sep Step Decay Noise Multiplier \sep Privacy Cost Gap \sep Fairness \sep Thinwall \sep  Global Scaling
%%PACS codes here, in the form: \PACS code \sep code

\end{keyword}

\end{frontmatter}

%\linenumbers


\section{Introduction}
Deep learning emerged as a powerful technology during the fourth industrial revolution~\cite{sarker2021deep}. Business intelligence, sentiment analysis, banking, healthcare~\cite{ardila2019end}, finance ~\cite{huang2020deep}, and many other fields employ deep learning to earn huge revenue and reduce human burden~\cite{healthcaredive, FinancialServices}. Regrettably, data, such as patient images~\cite{hassani2020deep}, used to train deep learning algorithms in specific industries mentioned above, are highly sensitive to privacy. Recent research shows that it is possible to extract sensitive information from deep learning models through different attacks~\cite{shokri2017membership, hu2022membership,truex2018towards, gong2016you, gong2018attribute,zhao2021feasibility, fredrikson2015model,wu2016methodology,chen2020improved, dwork2014dwork,dinur2003revealing}. Even more concerning, sensitive information cannot be protected using conventional methods like de-identification~\cite{near2021programming} and $k$-anonymity~\cite{sweeney2015only}. As most real-world applications often have unbalanced data, it is very important to consider fairness between groups in the data set. The existing investigation indicates that differential privacy (DP) can provide strong privacy guarantees for sensitive information ~\cite{dwork2011differential, dwork2006calibrating}. However, fairness and accuracy under differential privacy are exacerbated ~\cite{farrand2020neither}.

In deep learning, DP-SGD~\cite{abadi2016deep, chen2020understanding} is used more frequently as it obtains higher accuracy with a reasonable loss of privacy. Several methods have been proposed to improve the trade-off between privacy, utility, and fairness in DP-SGD. Zhang ~\textit{et al.}~\cite{zhang2021adaptive} developed a linear decay noise multiplier to decrease noise as training progresses. We observe that linear decaying of the noise multiplier has little or no advantage in improving the performance of the model. Bu ~\textit{et al.} ~\cite{bu2024automatic} introduced automatic clipping (Auto-S) and Yang ~\textit{et al.}  ~\cite{yang2022normalized} proposed normalized SGD (NSGD), using normalization instead of clipping to limit gradient sensitivity. However, Xia ~\textit{et al.}~\cite{xia2023differentially} highlighted that the above methods introduce a larger deviation between normalized and unnormalized average gradients. To decrease such a deviation, DP-PSAC ~\cite{xia2023differentially} used a non-monotonic adaptive weight function to limit the sensitivity of gradients. Nevertheless, none of these approaches successfully offered a solution to enhance fairness within the DP framework.

\begin{figure}
\centering
\begin{center}
\includegraphics[width=1.0\textwidth,height=0.4\textheight,draft=false]{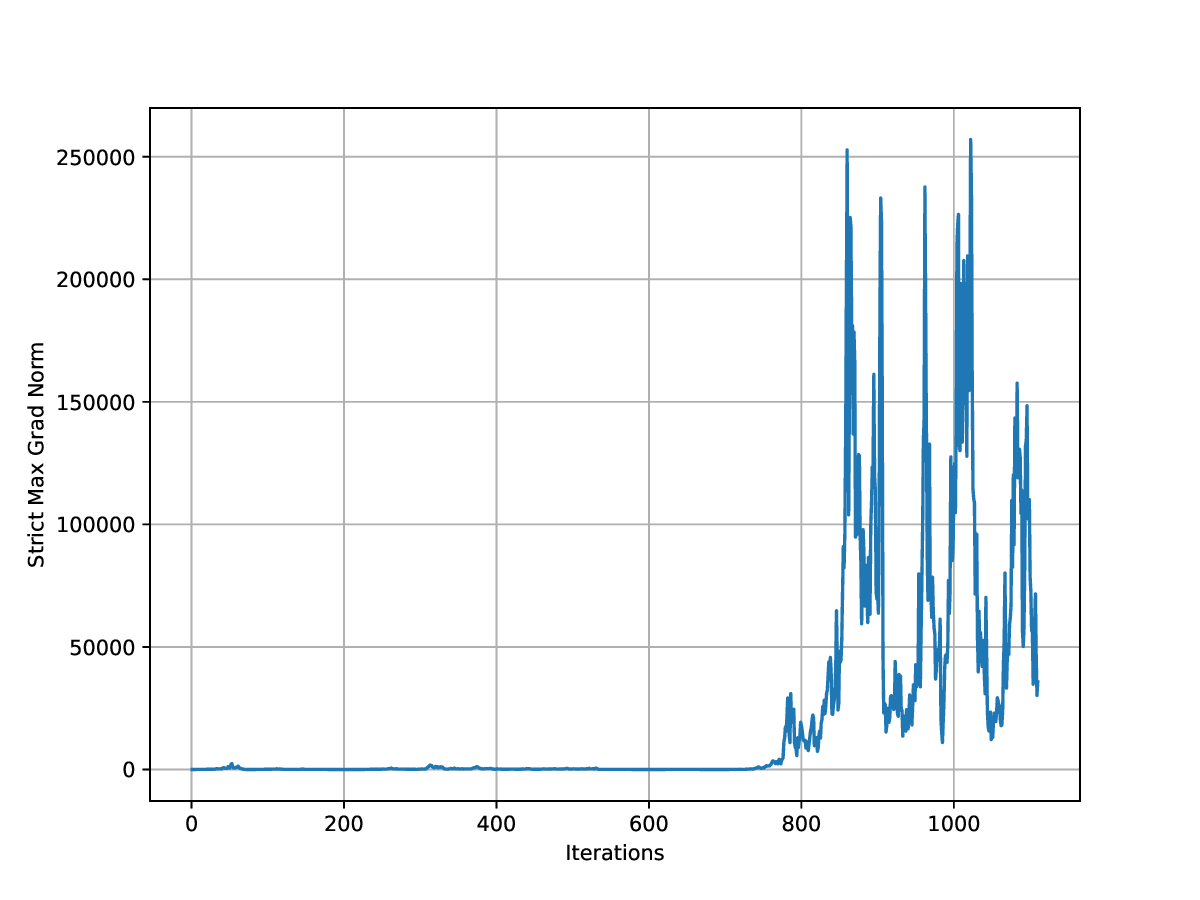}
\caption{Upper clipping threshold (strict max grad norm) during the training at every iteration for DP-SGD-Global-Adapt~\cite{esipova2022disparate} uing MNIST data. We use AdamW optimizer, OCL LR scheduler, and batch size of 64 for training and recorded the upper clipping threshold of DP-Global-Adapt after every iteration.} 
\label{strict max grad norm}
\end{center}
\end{figure}

In fairness, under Differential Privacy (DP), global scaling methods have gained prominence. These algorithms employ upper and lower clipping thresholds to scale the gradients. DP-Global ~\cite{bu2021convergence} scales all gradients that are below the upper clipping threshold (strict clipping bound) and discards those that exceed it, potentially leading to loss of information. DP-Global-Adapt ~\cite{esipova2022disparate} prevents information loss by clipping gradients that exceed the upper clipping threshold. Moreover, it uses a geometric update rule to dynamically adjust the upper clipping threshold so that all gradient norms are lower than the upper clipping threshold. The global sensitivity of both DP-Global and DP-Global-Adapt is equivalent to the lower clipping threshold. From Fig.~\ref{strict max grad norm}, we can observe that the upper clipping threshold of DP-Global-Adapt has increased exponentially at the end of the training. This behavior may hinder or slow the convergence of the model and require an additional privacy budget to converge. The exponential increase in the upper clipping threshold scales the gradients to extremely small magnitudes because the scaling factor is inversely proportional to the upper clipping threshold. To have better model convergence, we use a step-based decaying upper clipping threshold in DP-Global-Adapt-V2-S. When the gradient norm exceeds the upper clipping threshold, we use DP-PSAC's~\cite{xia2023differentially} clipping. DPSAC's clipping minimizes the discrepancy between the true batch average gradient and the model update. Furthermore, the integration of the proposed step decay noise multiplier improves the utility, fairness, privacy, and convergence of the model.

\begin{figure}
\centering
\begin{center}
\includegraphics[width=1.0\textwidth,height=0.4\textheight,draft=false]{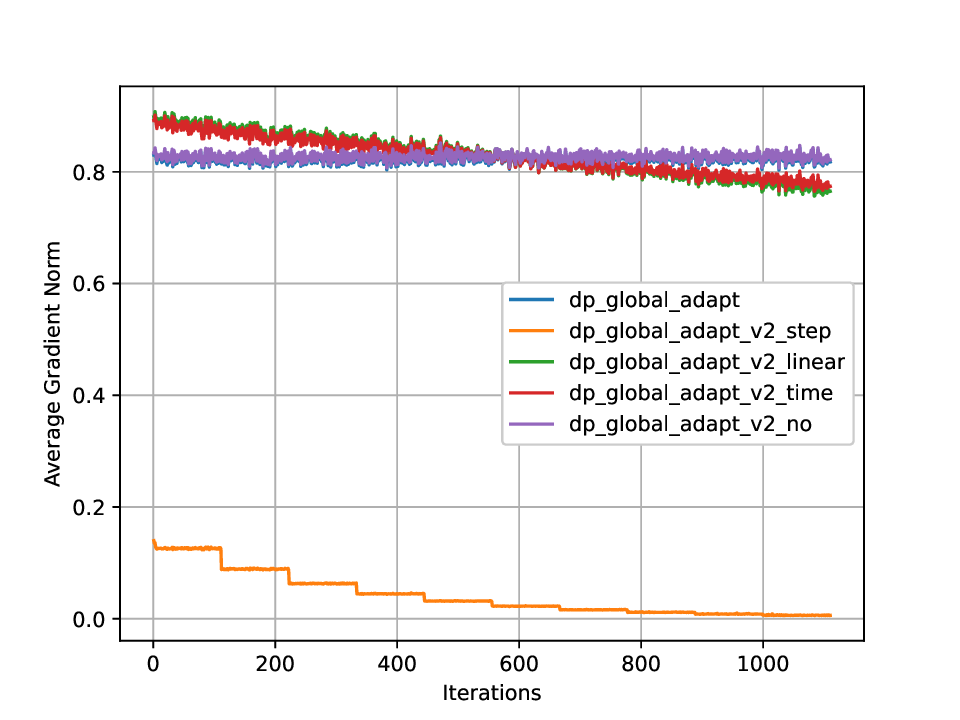}
\caption{Average gradient norm of  DP Global Adapt and all the versions of DP Global Adapt v2 during the training at every iteration. We use AdamW optimizer, OCL LR scheduler, and batch size of 64. During the mini-batch training, we record the gradient norm of every sample and then compute the average gradient in every iteration considering all the 64 samples gradient norm.}
\label{avg grad norm}
\end{center}
\end{figure}

As shown in Fig.~\ref{avg grad norm}, the average gradient norm for DP-Global-Adapt and DP-Global-Adapt-V2-no (where "no" indicates the absence of noise multiplier decay) remains constant during training, which affects the model's convergence. The remaining models that used noise multiplier decay showed a decrease in the average gradient norm through training. The step decay noise multiplier has a lower average gradient norm throughout the training. Linear and time-decay noise multipliers have a similar pattern of decreasing average gradient norm throughout the training. The point to be noticed from Fig.~\ref{avg grad norm} is that the average gradient norm during training is similar to the way that noise is added to the model. For example, DP-Global-Adapt-V2-step's decay of the average gradient norm resembles the step decay. Hence, employing the noise multiplier decay is essential for achieving a decreasing average gradient norm, which in turn facilitates faster model convergence. Our work introduces the following primary contributions: 

\begin{itemize}

\item We find that the latest DP-SGD-Global-Adapt~\cite{esipova2022disparate} method exhibits a convergence issue and the average gradient norm being the same throughout the training. To improve the convergence of the model and overall performance, we propose DP-SGD-Global-Adapt-V2-S. It uses a step-decaying upper clipping threshold and integrates the step-decay noise multiplier. Moreover, we have incorporated the DP-PSAC's~\cite{xia2023differentially} clipping when the gradient norm is higher than the upper clipping threshold. This helps reduce the deviation between the true gradient and the model update.

\item We propose the time and step noise multiplier decay mechanisms inspired by linear decay~\cite{zhang2021adaptive}. 
Decay of the noise multiplier reduces the noise multiplier after every epoch and helps to improve the accuracy of the model~\cite{zhang2021adaptive}. Moreover, we develop mathematical expressions to estimate the privacy budget using the tCDP accountant for the proposed noise multiplier decay variants.

\item We investigate all the noise multiplier decay mechanisms that affect DP models' accuracy, fairness, and privacy. We find that the step noise multiplier has better results in our experiments. So, we provide the reasons why the step noise multiplier has better results and explain how to choose the hyper-parameters for it.

\item DP-SGD-Global-Adapt-V2-S at the privacy budget ($\epsilon$) of 1, improves accuracy by 0.9795\%, 0.6786\%, and 4.0130\% in MNIST, CIFAR10, and CIFAR100, respectively. Moreover, it reduces the privacy cost gap ($\pi$) by 89.8332\% and 60.5541\% in unbalanced MNIST and Thinwall datasets, respectively. Furthermore, when evaluating the Thinwall dataset, we have used focal loss instead of cross-entropy loss. Focal loss was designed to address the data imbalance problem. 
 
\end{itemize} 

\section{Related Work} 
There are many studies related to DP-SGD. This section only compares the latest work with our DP-SGD-Global-Adapt-V2.

\textit{\textbf{Applications of Differential Privacy.}}
Integrating differential privacy (DP) into applications in e-Commerce~\cite{zhu2023informational}, environmental forecasting~\cite{zhu2024novel} and recommendation systems~\cite{cai2022deep} is essential to address the significant need to protect sensitive data against unauthorized exposure. In e-commerce, DP safeguards consumer insights used to personalize flash sale strategies, allowing companies to analyze imitation behaviors and purchase intentions without compromising individual privacy. This protection fosters consumer trust and aligns with ethical data use in marketing analytics. In environmental forecasting, DP preserves the confidentiality of sensitive national emissions data, improving model accuracy while supporting transparent decision-making processes with stakeholders. For recommendation systems, incorporating DP into advanced models such as DeepCGSR mitigates the risk of exposing personal preferences embedded in user-item ratings and reviews. This is particularly valuable given the high sensitivity of user sentiment and shopping behavior data in commercial contexts. By preserving privacy at the data processing level, DP not only ensures compliance with data protection standards but also contributes to the robustness of models used in diverse applications, supporting their secure deployment in privacy-sensitive domains.

\textit{\textbf{Differential Privacy.}}
In deep learning, DP-SGD is a popular scheme that uses gradient clipping and noise addition to protect sensitive information from individual data points at the expense of utility. To improve the trade-off between utility and privacy, the current literature has developed techniques to adaptively change the noise multiplier or clipping threshold. Zhang ~\textit{et al.}~\cite{zhang2021adaptive} proposed an adaptive DP-SGD that linearly decays the noise multiplier. Furthermore, they showed that adaptive DP-SGD achieved better performance than DP-SGD. According to Bu ~\textit{et al.} ~\cite{bu2024automatic} and Yang ~\textit{et al.} ~\cite{yang2022normalized}, normalization is effective in constraining the sensitivity of the gradient compared to clipping. So, Bu ~\textit{et al.} introduced Automatic Clipping (Auto-S)~\cite{bu2024automatic} and Yang ~\textit{et al.} proposed Normalized Stochastic Gradient Descent (NSGD). 
Furthermore, they demonstrated that when all gradients are normalized to the same magnitude, the learning rate and the clipping hyper-parameter can be linked, allowing the tuning of just one hyper-parameter. However, this method is known to exhibit significant variations between the normalized batch-averaged gradient and the unnormalized one when certain gradient norms in a batch are extremely small. Therefore, Xia ~\textit{ et al.} ~\cite{xia2023differentially} developed a Differentially Private Per Sample Adaptive Clipping algorithm (DP-PSAC). DP-PSAC uses a non-monotic adaptive weight function to clip the gradients and reduce the deviation between the update and the true batch-averaged gradient. 

\textit{\textbf{Fairness-aware Differential Privacy.}}
As deep learning is widely used in many highly regulated industries, data privacy and fairness must be carefully considered. Under DP, fairness has been shown to be exacerbated ~\cite{farrand2020neither}. To alleviate this, Xu ~\textit{ et al. } ~\cite{xu2021removing} proposed a DPSGD-F to offer equal privacy cost between groups and high utility. 
DPSGD-F adjusts the contribution of samples within a group based on the group clipping bias, thereby ensuring that differential privacy does not adversely affect group accuracy. DPSGD-F requires access to group labels, which can lead to privacy violations. 
DP-SGD Global~\cite{bu2021convergence} and DP-SGD Global-Adapt~\cite{esipova2022disparate} are popular gradient scaling algorithms to improve fairness that do not require access to the group labels. 
DP-SGD Global scales gradients with an $l_2$ norm less than or equal to a clipping threshold. If the gradients are larger than the clipping threshold, they are discarded. DP-SGD Global has two problems: (i) If the clipping threshold is set too high, no gradients are discarded, but the clipped gradients become smaller, making it harder for the algorithm to converge. (ii) If the clipping threshold is too low, most gradients are discarded, leading to information loss. DP-SGD-Global-Adapt is a modified version of DP-SGD-Global that allows the DP algorithm to achieve better fairness. It does this by clipping the gradients higher than the upper clipping threshold to have an $l_2$ norm equal to the lower clipping threshold to reduce information loss. It also adaptively changes the upper clipping threshold to be higher than all the per-sample gradients in a differentially private manner. 
The scaling factor used in both global methods is the lower clipping threshold over the upper clipping threshold. The DP-SGD Global-Adapt method uses a geometric update rule to adjust the upper clipping threshold, which increases significantly, particularly at the end of training. This exponential increase causes the gradients to scale down exponentially, which may impede convergence. 
Moreover, DP-SGD-Global-Adapt requires an additional privacy budget to modify the upper clipping threshold.

Previous studies focused on improving either the noise multiplier or the clipping threshold. Our proposed DP-SGD-Global-Adapt-V2 approach aims to optimize both the noise multiplier and the clipping threshold. To prevent an exponential increase in the upper clipping threshold, we utilize step decay to modify the upper clipping threshold, as gradients typically decrease as training progresses. When gradients exceed the upper clipping threshold, we use the clipping mechanism of DP-PSAC~\cite{xia2023differentially} to minimize the difference between the model update and the true batch average gradient. 
Furthermore, we have incorporated noise decay schedulers, including linear~\cite{zhang2021adaptive}, time, and step, so that the average gradient norm of the model decreases during model training.

\textit{\textbf{Privacy accountants.}} Privacy accountants calculate the loss of privacy incurred during each iteration of the DP training to calculate the total cost of privacy $(\epsilon, \delta)$. 
Dwork~\textit{et al}.~\cite{dwork2006our, dwork2009differential} offers a simple composition method that linearly combines the DP of various iterations, resulting in a greater loss of privacy. Dwork~\textit{et al.}~\cite{dwork2010boosting} defined an advanced composition theorem to tightly bind the cumulative privacy budget. 
Abadi~\textit{et al.}~\cite{abadi2016deep} showed that tighter estimates of total privacy loss could be obtained by tracking higher moments of privacy loss. 
Mironov~\textit{et al.}~\cite{mironov2017renyi} introduced an RDP based on Rényi divergence to track cumulative privacy loss throughout training. 
RDP underestimates the true cost of privacy. Dong~\textit{et al.}~\cite{dong2019gaussian} proposed $f$-DP to measure privacy cost from the point of view of hypothesis testing, with GDP as the main application. However, while GDP permits a tight composition, it is computationally difficult to determine the accurate composition of the Gaussian mechanism with subsampling amplification. 
Bun~\textit{et al.}~\cite{bun2018composable} proposed tCDP as an enhancement over CDP. 
tCDP supports privacy amplification, unlike CDP, and offers a method to increase accuracy exponentially. 
Recently, Gopi~\textit{et al.}~\cite{gopi2021numerical} have proposed numerical methods to determine the optimal composition of the DP mechanisms. 
It is difficult to calculate how much privacy the algorithm loses when the noise multiplier changes for each epoch, as in DP-SGD-Global-Adapt-V2. 
Therefore, we consider using tCDP to estimate the privacy budget of the proposed algorithm, as it provides a way to accommodate the changing noise multiplier.

\section{Background}
\label{BG}
This section discusses differential privacy and focal loss.

\subsection{Differential Privacy}
Differential privacy (DP)~\cite{dwork2008differential, hilton2002differential} is a method to preserve an individual's data while revealing aggregated information. DP is formally defined as follows:

\begin{definition}
A randomized function $\mathcal{F}: \mathcal{D} \rightarrow \mathcal{R}$ with a domain $\mathcal{D}$ and range $\mathcal{R}$ satisfies the differential privacy $(\epsilon, \delta) - $ if for any two data sets $d$, $\hat{d} \in \mathcal{D}$, differing with only a single data sample and for any subset of outputs O $\subseteq$ $\mathcal{R}$, we have 
  
\begin{equation}
    Pr[\mathcal{F}(d) \in O] \leq e^{\epsilon}Pr[\mathcal{F}(\hat{d}) \in O)] + \delta
\label{eq1}
 \end{equation}
\end{definition} 

One commonly used method to introduce randomness to a deterministic real-valued function $g: \mathcal{D} \rightarrow \mathcal{R}$ is by adding noise calibrated to the sensitivity $C$ of the function $g$. Sensitivity is the maximum absolute difference between the output of $g$ in any two neighboring data sets $d$, $\hat{d} \in \mathcal{D}$. 
In DP-SGD, the gradients are perturbed to provide a privacy guarantee for the deep neural network. 
The sensitivity is enforced by clipping the $L_{2}$-norm of the gradient. 

\begin{equation}
   C = max_{d,\hat{d}} || g(d) - g(\hat{d})||_{2}
\label{eq2}
\end{equation}

Most commonly, noise is drawn from the Gaussian distribution and added to the deterministic function as follows:

\begin{equation}
 \mathcal{F}(d) = g(d) + \mathcal{N}(0,\sigma^{2}I)
 \label{eq3}
\end{equation}

where $\mathcal{N}(0,\sigma^{2}I)$ is the Gaussian distribution with mean 0 and standard deviation $\sigma I$ and $\sigma$ is termed the noise multiplier.

The function $\mathcal{F}$ satisfies $(\epsilon, \delta)-$ DP, where $\delta \in (0,1)$ and $\sigma \geq \frac{\sqrt{2ln(1.25)/}\Delta C}{\epsilon}$ is a noise multiplier.

\begin{definition}
    (tCDP). For all $\alpha \in (1,\omega)$, a randomized algorithm $\mathcal{A}$ is $(\rho, \omega) -$ tCDP if for any neighboring data sets $d$ and $\hat{d}$, and for all $\alpha > 1$, we have:  

\begin{equation}
     D_{\alpha}(\mathcal{A}(d)||\mathcal{A}(\hat{d})) \leq \rho\alpha
\end{equation}
   
\end{definition}

where $D_{\alpha}(\cdot||\cdot)$ is the Rényi divergence of order $\alpha$. 

Given two distributions $\mu$ and $\nu$ on a Banach space (Z,$|| \cdot ||$), the Rényi divergence is calculated as follows:

\begin{definition}
Rényi divergence~\cite{renyi1961measures}: Let $1 < \alpha < \infty$ and $\mu,\nu$ be measures with $\mu \ll \nu$. The Rényi divergence of orders $\alpha$ between $\mu$ and $\nu$ is defined as:  

    \begin{equation}
        D_{\alpha}(\mu||\nu) \doteq \frac{1}{\alpha - 1}ln\int(\frac{\mu(z)}{\nu(z)})^{\alpha}\nu(z)dz.
    \end{equation}
    
\end{definition}

Here we follow the convention $\frac{0}{0} = 0$. If $\mu \not \ll \nu$, we define the Rényi divergence as $\infty.$ The Rényi divergence of orders $\alpha = 1, \infty$ is defined by continuity. 

In this work, we mainly use the following properties of tCDP, as demonstrated in \cite{bun2018composable}: 

\begin{lemma}
\label{lemma1}
The Gaussian mechanism satisfies $(\frac{C^{2}}{2\sigma^{2}},\infty)-$ tCDP.
\end{lemma}

\begin{lemma}
\label{lemma2}
   If randomized functions $\mathcal{F}_{1}$ and $\mathcal{F}_{2}$ satisfy $(\rho_{1},\omega_{1})$-tCDP and $(\rho_{2},\omega_{2})$-tCDP, their composition defined as ($\mathcal{F}_{1} \circ \mathcal{F}_{2}$) is $(\rho_{1} + \rho_{2}, min(\omega_{1},\omega_{2}))-$tCDP.
\end{lemma}

\begin{lemma}
\label{lemma3}
    If a randomized function $\mathcal{F}$ satisfies $(\rho, \omega)-$tCDP, then for any 
    $\delta \geq \nicefrac{1}{exp((\omega -1)^{2}}\rho)$,  
    $\mathcal{F}$ satisfies $(\rho +2\sqrt{\rho ln(1/\delta)}, \delta)-$ differential privacy.
\end{lemma}

\begin{lemma}
\label{lemma4}
     If a randomized function $\mathcal{F}$ satisfies $(\rho, \omega)-$tCDP, then for any $n$-element data set $D$, computing on uniformly random $hn$ entries ensures $(13h^{2}\rho, log(1/h)/(4\rho))-$tCDP, with $\rho, h \in(0,0.1]$, $log(1/h) \geq 3\rho(2+ log(1/\rho))$ and $\omega \geq log(1/h)/(2\rho)$. 
\end{lemma}

The lemma \ref{lemma1} provides a relation between the Gaussian mechanism and the tCDP privacy accountant. 
The lemma \ref{lemma2} describes the composition property of two randomized functions under tCDP. 
The lemma \ref{lemma3} provides a way to convert the privacy budget of the tCDP accountant to the standard $(\epsilon, \delta)-$ DP. 
The lemma \ref{lemma4} illustrates privacy amplification through random sampling using tCDP. We derive the mathematical expression to compute tCDP for our proposed algorithm using these lemmas as a basis in Section~\hyperref[8]{8}

\subsection{Focal Loss} 

Focal loss modifies the cross-entropy loss function to handle classification tasks, especially in situations with imbalanced datasets and for binary classification. It incorporates a modulating factor of ${(1-p_{t})}^{\gamma}$ in the cross-entropy loss, where $p_t$ represents the predicted probability for the positive class and $\gamma$ is a hyperparameter. This factor decreases the loss for examples that are easy to classify (when $p_t$ is high), which generally belong to the majority class. Using ${(1-p_{t})}$ for examples where $p_t$ is low, often of the minority class, the model is encouraged to focus on those instances that are harder to classify, thereby enhancing the classification performance in unbalanced datasets. The focal loss also has a hyperparameter $\alpha$ that controls the weight of the modulating factor. A higher alpha gives more weight to the minority class.

\begin{equation}
    FL(p_t) = -\alpha (1 - p_t)^{\gamma} \log(p_t)
\end{equation}

\section{Methodology} 

In this section, we discuss the DP-SGD-Global-Adapt-V2, different types of noise multiplier decay schedulers, and how to compute the total privacy budget using tCDP for all the DP-SGD-Global-Adapt-V2 variants.

\subsection{DP-SGD-Global-Adapt-V2} 

DP-SGD-Global-Adapt-V2 takes the dataset $D$, a lower clipping bound $c_0$, an upper clipping bound (strict clipping bound) $z_0$, a noise multiplier decay mechanism $\sigma_e^2$, a clipping decay mechanism $z_e$, and some other parameters as input. Then, it initializes the model parameters, noise multiplier, and strict clipping bound. The algorithm runs $T$ iterations, where $T = \frac{E}{q}$. $E$ represents the total number of training epochs. $q=b/n$ is the sampling rate, where $b$ is the batch size and $n$ represents the number of training examples in the data set. In each iteration, the algorithm first takes the batch of samples $b$ from the data set $D$ according to the sampling rate of poisson $q$. Then, for every sample in the batch, the algorithm computes the gradient. Next, it computes the scaling factor $\gamma_i$. If the $l_2$ norm of the sample gradient $g_i$ is less than the strict clipping bound $z_e$, then the scaling factor is $\frac{c_0}{z_e}$, where $c_0 < z_e$. 

Otherwise, the scaling factor is $\frac{c_{0}}{(||g_i|| + \frac{w}{||g_i|| + w})}$. Here, $w$ is a constant that is chosen before the training. Then, the scaled gradient is calculated as $\bar{g_b} = \gamma_i \cdot g_i$. 
Next, the noise multiplier for the current epoch, $\sigma_e^2$, is calculated according to the decay type $m$. The noise multiplier is computed after every epoch (not after every iteration). 
Then Gaussian noise with a mean of zero and a variance of $\sigma_e^2$ is added to the batch average of the scaled gradients. 
Finally, the model is updated using gradient descent, and the strict clipping bound $z_e$ is updated for the next epoch according to the step decay based clipping mechanism. 
After running through all the iterations, the final model is obtained and can be used for inference. 
DP-Global ~\cite{bu2021convergence}, DP-Global-Adapt~\cite{esipova2022disparate} and DP-SGD-Global-Adapt-V2 would guarantee a global bounded sensitivity of $c_0$, since all gradient norms are bounded to $c_0$ ~\cite{esipova2022disparate}. 

\begin{algorithm}[!tp]
\caption{DPSGD-Global-Adapt-v2}\label{alg:Global-Adapt-V2}

\KwData{Dataset $D$, sampling rate $q$, clipping bound $c_{0}$, strict clipping bound $z_{0}$, epochs $E$, decay rate $R$, epoch drop rate $K$, noise multiplier decay mechanism $\sigma_e^2 = F(e, R, K, \sigma_0, m)$, learning rate $\eta_{t}$, batch size $B$, noise multiplier decay type $m$, clipping decay mechanism $z_{e}$ = $G(e,R,K,z_{0})$, iteration $T = \frac{E}{q}$.}

\textbf{Initialize $\theta_{0}$, $\sigma_{0}$, $z_0$.} \\

\For{$t$ in $0, 1, ..., T-1$}{
    $B$ $\leftarrow$ Poisson sample of $D$ with sampling rate $q$.
    
    \For{($x_i$, $y_i$) in $B$}{
        $g_i \leftarrow  \bigtriangledown_{\theta_{l}}(S_{\theta_{t}}(x_i), y_i)$\\
        $\gamma_i \leftarrow
        \begin{cases} 
        \frac{c_0}{z_e}, & \text{if } ||g_i|| \leq z_{e}, \ e = \lfloor q\cdot t \rfloor \\
        \frac{c_{0}}{(||g_i|| + \frac{w}{||g_i|| + w})}, & \text{if } ||g_i|| \geq z_{e}
        \end{cases}$\\
        $\bar{g_i} \leftarrow \gamma_{i} \cdot g_i$ \\
    }
    $\sigma_{e}^2 \leftarrow F(e, R, K, \sigma_0, m)$, $e = \lfloor q\cdot t \rfloor$\\
    $\tilde{g_B} \leftarrow \frac{1}{|B|}(\Sigma_{i \in B} \bar{g_i} + \mathcal{N}(0, \sigma_{e}^2 \cdot \mathbf{I}))$\\
    $\theta_{t+1} \leftarrow \theta_{t} - \eta_t \cdot \tilde{g_B}$\\
    $z_{e} \leftarrow G(e,R,K,z_{0})$
}
\KwResult{$\theta_T$}

\end{algorithm}

\begin{table}[!hbtb]
\centering
\caption{Types of noise multiplier decay mechanisms.}
\label{tab1}
\begin{tabular}{@{}ll@{}}
\toprule \small
Decay type & Mathematical expression \\ \midrule
Linear decay ~\cite{zhang2021adaptive}     & $\sigma_e^2 = \sigma_0^2  R^e$       \\
Time decay & $\sigma_e^2 = \frac{\sigma_0^2 }{1+Re}$     \\
Step decay & $\sigma_e^2 = \sigma_0^2  R^{\lfloor e/K \rfloor}$          \\             
 \bottomrule
\end{tabular}
\end{table}

\subsection{The decay schedulers}  

As training progresses, the gradients should decrease for DP-SGD~\cite{fang2022improved}. The noise multiplier is the same throughout the training in DP-SGD, DP-PSAC, DP-F, DP-Global, DP-Global-Adapt, and Auto-S. 
In that case, it is possible that the noise will overpower the gradients, especially in later training iterations when the gradients are much smaller, leading to meaningless model predictions. Moreover, it is necessary to decrease the noise multiplier through training to improve the utility~\cite{zhang2021adaptive}. Therefore, Zhang ~\textit{et al.}~\cite{zhang2021adaptive} used a linear decay noise multiplier to minimize the negative impact of the addition of the same amount of noise during training. We build upon their work and propose two more decaying mechanisms: step decay and time decay, inspired by learning rate schedulers used in non-private settings. The various noise multiplier decay techniques examined in this paper are illustrated in Table \ref{tab1}. 
In Table \ref{tab1}, $\sigma_{0}$ is the initial noise multiplier; $e$ is the epoch number; $R$ is the decay rate; $K$ is the epoch drop rate; and $E$ is the total number of training epochs. 
Linear and time decay adjust after each epoch, while step decay modifies after each step size. Linear and time decay change the noise multiplier gradually, whereas step decay changes the noise multiplier rapidly. 
Moreover, we use the step-based clipping decay mechanism to update the strict clipping bound, which is $z_e = z_0 R^{\lfloor e/K \rfloor}$. 

\subsection{Cumulative privacy budget for DP-SGD-Global-Adapt-V2 variants}

In this work, we mainly use the lemma ~\ref{lemma1}- ~\ref{lemma4} of tCDP, as demonstrated in \cite{bun2018composable} and presented in the ~\hyperref[BG]{Background} Section. 

\begin{table}[!hbtb]
\centering
\caption{Total privacy budget for different noise multiplier decay mechanisms.}
\label{tab-9}
\begin{tabular}{@{}lll@{}}
\toprule \small
 Decay type & $\rho_{total}$ & $\omega_{total}$ \\ \midrule
 Linear     & $\frac{13(b/n)^2 C^2 (1 - R^E)}{2\sigma_0^2(R^{E-1} - R^E)}$ & $\frac{\log(n/b)\sigma_0^2 R^{E-1}}{2C^2}$ \\
 Time & $\frac{13(b/n)^2 C^2 (2E + R(E)(E-1))}{4\sigma_0^2}$ & $\frac{\log(n/b)\sigma_0^2}{2C^2(1 + R(E-1))}$\\
Step & $\frac{13(b/n)^2 C^2 K (1 - R^P)}{2\sigma_0^2 (R^{P-1} - R^P)}$ & $\frac{\log(n/b)\sigma_0^2 R^{P-1}}{2C^2 K}$ \\ 
No & $\frac{13(b/n)^2 C^2 E}{2\sigma_0^2}$ & $\frac{\log(n/b)\sigma_0^2}{2C^2}$ \\ \bottomrule
\end{tabular}
\end{table}

To estimate the cumulative privacy loss of the proposed algorithm, we use the composition theorem of tCDP. tCDP was created to support more computations and offer a sharper and tighter analysis of privacy loss than the strong composition theorem of $(\epsilon,\delta)$-DP. We provide mathematical expressions to compute the cumulative privacy budget for no decay, time decay, and step decay, and present the total privacy budget for linear decay~\cite{zhang2021adaptive}. 
Table \ref{tab-9} provides the mathematical expressions for computing the privacy budget for all variants of DP-SGD-Global-Adapt-V2, including when no noise multiplier decay is used. In Table \ref{tab-9}, $\sigma_{0}$ is the initial noise multiplier; $E$ is the total number of training epochs; $R$ is the decay rate; $n$ is the total number of training samples; $b$ is the batch size; $K$ is the epoch drop rate; $C$ is the sensitivity of the gradients; and the ratio of the total number of training epochs to the epoch drop rate is $P=E/K$. To derive the final expression for the step decay, we simplified the step decay from $\sigma_e^2 = \sigma_0^2 R^{\lfloor e/K \rfloor}$, $R=0.5$, and $K=10$ to $\sigma_p^2 = D \sigma_0^2 R^p$, where $p$ ranges from 0 to $P-1$, and we assume that $E$ is divisible by $K$. After obtaining $\rho_{total}$ and $\omega_{total}$, we can apply the lemma~\ref{lemma3} to calculate the corresponding privacy parameters. Specifically, $\epsilon$ should be set to $(\rho_{total} +2\sqrt{\rho_{total}\ln(1/\delta)})$, where $\delta$ is a predetermined fixed value that represents the probability of failure. We provide detailed derivations of the total privacy budget of DP-SGD-Global-Adapt-V2, DP-SGD-Global-Adapt-V2-S, DP-SGD-Global-Adapt-V2-L, and DP-SGD-Global-Adapt-V2-T in Section~\hyperref[8]{8}. 

\section{Experiments} 

In this paper, we use the DP and DP-SGD interchangeably. We want to emphasize that the lower the privacy budget ($\epsilon$), the higher the privacy and the less vulnerable the model is to inference or any other attacks. We explain all the notation of the project in Table~\ref{tab_not}. 
In this section, we explain the datasets and describe the findings on privacy, utility, and fairness. We then compare the various types of DP-SGD-Global-Adapt-V2 and provide the noise multiplier decay scheduler analysis. Lastly, we demonstrate how to select the hyperparameters for the step decay noise multiplier and analyze the model training hyperparameters. 

\begin{table}[!hbtb]
\centering
\caption{Explanation of all the notation used in this work.}
\label{tab_not}
\begin{tabular}{@{}ll@{}}
\toprule
Notation                   & Explanation                                    \\ \midrule
$S$                        & Model                                        \\
$\theta_t$                 & Parameters of the model in $t^{th}$ iteration \\
$x_i$                      & $i^{th}$ data sample \\
$y_i$                      & $i^{th}$ target sample \\
%$\bigtriangledown$         & Model gradient \\
$b$                        & Batch size                                       \\
$\sigma_{0}$ & Initial noise multiplier                                      \\
$e$                        & Current epoch number                            \\
$E$                        & Total number of training epochs                  \\
$R$                        & Decay rate                                      \\
$q$                        & Sampling rate                                   \\
$n$                        & Total number of samples in the data set.       \\
$T$                        & Total number of training iterations             \\
$t$                        & Current iteration number                        \\
$\eta_{t}$                 & Learning rate at $t^{th}$ iteration              \\
$K$                        & Epoch drop rate                                 \\
$F$                        & noise multiplier decay mechanism        \\
$G$                        & upper clipping threshold step decay mechanism \\
$m$                        & noise multiplier decay type \\
$C$                        & The sensitivity of gradient                \\
$z_e$                      & Upper clipping threshold at epoch $e$ \\
$\gamma_i$                 & scaling factor for sample $i$ \\
$\rho$, $\omega$           & Privacy parameters of tCDP                      \\
$\rho_e$, $\omega_e$       & The $\rho$, $\omega$ at $e^{th}$ epoch.             \\
$\rho_{total}$,  $\omega_{total}$ & Total privacy budget parameters of tCDP          \\
$\bar{g_i}$           & scaled gradient of the $i^{th}$ training data  \\ & sample in a batch \\
$\tilde{g_B}$                & Average of noisy and scaled gradients \\
$\mathcal{N}(0, \sigma^2\mathbf{I})$ & The gaussian distribution with mean 0 \\
&and standard deviation $\sigma$\\
$P$                        & $P = \frac{E}{K}$  \\ 
$p$                        & An integer that ranges from 0 to $P$-1  \\       \bottomrule

\end{tabular}
\end{table}

\textbf{MNIST.} MNIST data set~\cite{lecun1998gradient} consists of grayscale images of digits ranging from 0 to 9 with dimensions of $28 \times 28$ pixels. The training set comprises 60,000 images, while the test set contains 10,000 images. For MNIST, we used a model built with two convolutional layers with 20 and 50 channels, respectively, with a kernel size of $5 \times 5$. On top of that, it has a two-layer classification layer with 500 hidden units. Moreover, it has ReLU activation after every layer and maxpool2d after every convolutional layer.

\textbf{CIFAR10.} CIFAR-10 data set~\cite{krizhevsky2009learning} consists of 60,000 color (RGB 3-channel) images with dimensions of $32 \times 32$ pixels. It includes 6,000 images per class, spanning across 10 classes. The data set is divided into 50,000 training images and 10,000 test images. The CIFAR10 is fine-tuned using a pre-trained NF-Net-F0~\cite{brock2021high} model. We resize the training images to $192 \times 192$ and the test images to $256 \times 256$. In the fine-tuning process, we reinitialize only the final classification layer. 

\textbf{CIFAR100.} CIFAR-100 data set~\cite{krizhevsky2009learning} consists of 60,000 color images divided into 100 classes, with 600 images per class. Like CIFAR-10, the CIFAR-100 data set also includes 50,000 training images and 10,000 test images. The CIFAR100 is fine-tuned using a pre-trained NF-Net-F1~\cite{brock2021high} model. We resize the training images to $224 \times 224$ and the test images to $320 \times 320$. In the fine-tuning process, we reinitialize only the final classification layer.

During experiments, we run the above three datasets for 100 training epochs with a batch size of 64, a one-cycle learning rate scheduler with an initial learning rate of 1e-4, and an Adamw optimizer with a weight decay of 1e-3. We chose these hyperparameters because of their best performance as shown in Section~\ref{analysis hp}. We use MNIST, CIFAR10, and CIFAR100 to evaluate utility. We use the unbalanced MNIST and the real-world application dataset, Thinwall, to evaluate fairness.

\textbf{Unbalanced MNIST.}We create an artificially unbalanced MNIST training data set where class 8 only constitutes about 1\% of the data on average. The test data set is the same as the MNIST data set. We use a two-layer CNN model consisting of 16 and 32 channels, respectively, with a kernel size of $3 \times 3$. We use instance normalization, a SeLU activation layer, and a maxpool2d layer. On top of that, there is the single-layer classification layer.

\textbf{Thinwall.} Thinwall dataset ~\cite{zamiela2023thermal} is gathered from the additive manufacturing process (AM), which includes thermal images of melt pools. In particular in the course of producing Ti-6Al-4V samples, these images were obtained using a coaxial, dual-wavelength pyrometer camera integrated into the OPTOMEC LENS 750 system ~\cite{khanzadeh2019situ, esfahani2022situ}. These thermal images, stored as comma-separated value (CSV) files, encompass temperature readings for each pixel in the field of view. Subsequently, X-ray computer tomography (XCT) was used to inspect the internal quality of the manufactured parts, revealing internal defects characterized by porosity. The classification of thermal images into healthy and anomalous was achieved through a manual matching process of melt pool images with corresponding XCT images, resulting in a significant disparity in the number of healthy and anomalous instances in the printed samples. 

The data set demonstrates a significant class imbalance due to the high stability of the AM process, where anomalies are rare events. This is quite common in anomaly detection for various AM processes~\cite{tian2020physics, bappy2022situ, ye2021situ, seifi2019layer}. During these AM, the process is in a healthy state for a much longer time compared to anomalous states, resulting in the majority of process thermal images being labeled as healthy. This leads to a skewed distribution of labels in the dataset. Although anomalies in AM processes are rare, they greatly affect the mechanical properties and functionality of the final product~\cite{al2020effects, sola2019microstructural, sanaei2021defects}. Therefore, it becomes crucial to accurately detect minority class samples (i.e., anomalous instances), as these anomalies hold critical insights into potential defects or irregularities within the manufactured parts. Therefore, ensuring the correct identification of such minority samples is vital for the overall reliability and efficacy of AM process monitoring and anomaly detection.

The Thinwall data set consists of 1494 nonporous (healthy) melt pool images and 70 porous (anomalous) melt pool images. The $752 \times 480$ resolution pyrometer images are cropped to the $200 \times 200$ resolution centered around the melt pool. We have created the train and test sets by dividing the data set into 75\% and 25\%, respectively. We have created a smaller version of ShuffleNet~\cite{ma2018shufflenet} to train a model using the Thinwall data set. The model has a convolutional layer with 16 channels, followed by ShuffleNet building blocks with 32 and 64 channels. The model consists of a kernel size of $3 \times 3$, the group normalization layer with 4 groups, the SeLU activation layer, and the maxpool2d layer. Finally, there is a classification layer. We explain the importance of differential privacy in additive manufacturing in Section~\ref{B}.

We employ the AdamW optimizer with a learning rate of 1e-3 to train both unbalanced MNIST and Thinwall datasets. The batch size for unbalanced MNIST is set to 64, while for Thinwall, it is set to 32 because the number of data samples is lower compared to others. For training DP-SGD-Global-Adapt-V2, we apply a strict clipping bound of 3. In the case of DP-Global, a strict clipping bound of 100 is used, and for DP-Global-Adapt, a strict clipping bound of 10 is utilized. The lower clipping bound for all algorithms is consistently set at 1.

\textit{\textbf{Implementation details.}}

We develop and implement codes for our experiments using PyTorch~\cite{Torch} and Opacus~\cite{yousefpour2021opacus}. We conduct all the experiments on a server equipped with an Intel Core i9-10980XE CPU, 251 GB of memory, and four Nvidia Quadro RTX 8000 GPUs, running Ubuntu 18.04 OS. For the implementation of DP-Global, DP-Global-Adapt, DP-SGD, and DP-F, we used the code provided by Xia ~\textit{et al.} ~\cite{xia2023differentially}. To implement DP-PSAC and Auto-S, we make changes to the Opacus optimizer according to the code given in the respective papers. We will release the codes for our proposed algorithm once the paper is accepted. 

\subsection{Results on privacy and utility}

In this section, we compare our proposed algorithm, DP-SGD-Global-Adapt-V2-S, against DP-PSAC, Auto-S, DP-SGD, DP-SGD-Global and DP-SGD-Global-Adapt. The "S" in DP-SGD Global Adapt-V2-S means that the algorithm uses a step decay noise multiplier during training. This section of experiments is focused only on improving privacy and utility and excludes fairness. Tables~\ref{tab_1}-\ref{tab_3} represent the accuracy of all DP methods in MNIST, CIFAR10, and CIFAR100 data sets, respectively. We evaluated every DP algorithm for five privacy budgets ($\epsilon$): 1, 3, 5, 8, and 10. From Tables~\ref{tab_1}-\ref{tab_3}, we make the following observations: First, on all the datasets considered, as the privacy budget increases, the accuracy of the model also increases. Secondly, the advanced methods DP-PSAC, Auto-S, DP-SGD-Global, and DP-SGD-Global-Adapt have lower performance compared to DP-SGD in some cases. To illustrate, consider Table~\ref{tab_1}, Auto-s, DP-PSAC, DP-SGD-Global, and DP-SGD-Global-Adapt obtain 97.99\%, 97.96\%, 93.98\%, and 97.71\% accuracy, while DP-SGD has 98.02\% at a privacy budget of 1. DP-SGD-Global exhibits the lowest performance among all the algorithms because of the information loss that occurs when gradients with norms exceeding the upper clipping threshold are discarded. However, the other existing works have the same or better performance as that of DP-SGD in some other cases. For example, DP-SGD and Auto-s have an accuracy of 94.30\% and 94.76\% at the privacy budget of 1 and 3, respectively, in the CIFAR10 data set. While DP-SGD-Global-Adapt has an accuracy of 95.07\%  and DP-SGD has 95.04\% accuracy at the privacy budget of 10. Finally, we can see that the DP-SGD-Global-Adapt-V2-S consistently outperforms all existing work in the three datasets considered. For example, with a privacy budget of 10, the DP-SGD-Global-Adapt-V2-S has an accuracy of 99.26\%, 95.24\%, and 80.30\%, while the best accuracy obtained from existing work is 99.23\%, 95.07\%, and 79.88\%  for MNIST, CIFAR10, and CIFAR100, respectively. The reason for the high performance of our method is the optimization of the noise multiplier and clipping mechanism. More importantly, the accuracy of DP-SGD-Global-Adapt-V2-S at a privacy budget of 10 is very close to that of Non-DP. In CIFAR10, the non-DP accuracy is 95.29\%, and the proposed algorithm got an accuracy of 95.24\% at the privacy budget of 10.

\begin{table}[htbp]
\centering
\caption{Comparision of Accuracy of existing DP-SGD algorithms against the Proposed Algorithm using MNIST data set.}
\label{tab_1}
\begin{tabular}{@{}lllllll@{}}
\toprule
$\epsilon$  & DP-SGD & Auto-S & DP-PSAC & DP-SGD-Global & DP-SGD-Global-Adapt & DP-SGD-Global-Adapt-V2-S \\ \midrule
1  & 98.02\%  & 97.99\%     & 97.96\%   & 93.98\%         & 97.71\%               & 98.98\%                  \\
3  & 98.77\%  & 98.77\%     & 98.74\%   & 96.27\%         & 98.69\%               & 99.21\%                  \\
5  & 98.98\%  & 99.03\%     & 99.02\%   & 97.08\%         & 98.94\%               & 99.23\%                  \\
8  & 99.12\%  & 99.11\%     & 99.11\%   & 97.65\%         & 99.12\%               & 99.25\%                  \\
10 & 99.16\%  & 99.23\%     & 99.16\%   & 97.91\%         & 99.20\%               & 99.26\%                 \\ \bottomrule
\end{tabular}
\end{table}

\begin{table}[]
\centering
\caption{Comparision of Accuracy of existing DP-SGD algorithms against the Proposed Algorithm using CIFAR10 data set.}
\label{tab_2}
\begin{tabular}{@{}lllllll@{}}
\toprule
$\epsilon$  & DP-SGD & Auto-S & DP-PSAC & DP-SGD-Global & DP-SGD-Global-Adapt & DP-SGD-Global-Adapt-V2-S \\ \midrule
1  & 94.30\%  & 94.30\%     & 94.31\%   & 92.29\%         & 93.82\%               & 94.95\%                  \\
3  & 94.76\%  & 94.76\%     & 94.76\%   & 93.50\%         & 94.59\%               & 95.11\%                  \\
5  & 94.93\%  & 94.94\%     & 94.94\%   & 93.92\%         & 94.86\%               & 95.18\%                  \\
8  & 94.96\%  & 94.96\%     & 94.96\%   & 94.17\%         & 95.03\%               & 95.22\%                  \\
10 & 95.04\%  & 95.04\%     & 95.04\%   & 94.34\%         & 95.07\%               & 95.24\%         \\        \bottomrule
\end{tabular}
\end{table}

\begin{table}[]
\centering
\caption{Comparision of Accuracy of existing DP-SGD algorithms against the Proposed Algorithm using CIFAR100 data set.}
\label{tab_3}
\begin{tabular}{@{}lllllll@{}}
\toprule
$\epsilon$  & DP-SGD & Auto-S & DP-PSAC & DP-SGD-Global & DP-SGD-Global-Adapt & DP-SGD-Global-Adapt-V2-S \\ \midrule
1  & 76.75\%  & 76.75\%     & 76.75\%   & 67.60\%         & 70.56\%               & 79.83\%                 \\
3  & 78.63\%  & 78.63\%     & 78.63\%   & 72.97\%         & 75.16\%               & 80.06\%                  \\
5  & 79.40\%   & 79.40\%     & 79.40\%   & 74.76\%         & 76.43\%               & 80.21\%                  \\
8  & 79.74\%  & 79.74\%     & 79.74\%   & 76.14\%         & 77.25\%               & 80.27\%                  \\
10 & 79.87\%  & 79.88\%     & 79.88\%   & 76.56\%         & 77.63\%               & 80.30\%          \\        \bottomrule
\end{tabular}
\end{table}

\subsection{Results on privacy and fairness} 
This section discusses the experiments that focus on improving fairness under DP. We compare our work with those of DP-SGD, DP-F, DP-SGD-Global, and DP-SGD-Global-Adapt. To evaluate fairness, we use the ROC-AUC score, as it is robust to class imbalance and independent of the decision threshold. We also use accuracy parity, as in Bagdasaryan ~\textit{et al.}~\cite{bagdasaryan2019differential}. 

\textit{\textbf{Accuracy parity:}} Accuracy parity is defined as the difference in classification accuracy between protected groups after adding privacy. The protected group is the classification label in our experiments. We represent the subgroup of the data that contains data samples from the group $m$ as $D_m =  {(x_j, a_j, y_j) \in D|a_j = m}$. 
A private model has an accuracy parity for the subgroup $D_m$ as

\begin{equation}
   \pi_m = \pi(\theta, D_m) = acc(\theta^*; D_m) - E_{\tilde\theta}[acc(\tilde\theta;D_m)]
\end{equation}

Where the expectation is over the randomness involved in acquiring private model parameters. We use the privacy cost gap $\pi_{a,b} = |\pi_a - \pi_b|$ to measure fairness. Here, a and b are two different groups of a given dataset. Tables~\ref{tab_4}-\ref{tab_8} represent performance and fairness results (accuracy, AUC, group accuracy, and privacy cost gap) using DP-SGD, DP-F, DP-Global, DP-Global-Adapt, and DP-Global-Adapt-V2-S, respectively. In unbalanced MNIST, we measure the fairness of groups 2 and 8, similar to Esipova ~\textit{et al.}~\cite{esipova2022disparate}. However, accuracy and AUC are measured across all groups. In the Thinwall data set, the two groups are porous (defective) and non-porous (healthy). 
In Tables~\ref{tab_4}-\ref{tab_8}, we simply denote the privacy cost gap as $\pi$. We run experiments for five privacy budgets: 1, 3, 5, 8, and 10. From Tables ~\ref{tab_4}-\ref{tab_8}, we note the following points: Before detailed observations are made, it is important to note that a smaller privacy cost gap indicates better fairness of the algorithm. Firstly, in some cases, the privacy cost gap is very high for the DP-Global and DP-Global-Adapt methods. To illustrate, consider the unbalanced MNIST with a privacy budget of 1. The privacy cost gap is 92.7301\% and 94.3734\% for the DP-Global and DP-Global-Adapt methods, respectively. The high privacy cost gap is due to the inability of the algorithm to detect at least one sample from group 8 (the minority class). Another observation is that the DP-SGD has better fairness compared to the algorithms that are designed to have better fairness, such as DP-F, DP-Global and DP-Global-Adapt. At the privacy budget of 10, the DP-SGD has a privacy cost gap of 1.4373\%, while the DP-F, DP-Global, and DP-Global-Adapt have 2.1529\%, 16.8493\%, and 3.8045\%, respectively. Another interesting point is that the DP-Global-Adapt-V2 algorithm has better accuracy in identifying porous images than the non-DP algorithm. The non-DP algorithm has an accuracy of 81.82\%, while the DP-Global-Adapt-V2-S has an accuracy of 86.3636\% in identifying porous images at a privacy budget of 10. DP-Global-Adapt-V2-S consistently outperforms existing algorithms by a significant amount. To illustrate, consider Thinwall at a privacy budget of 1, the privacy cost gap for DP-Global-Adapt-V2-S is 27.0018\%, while the next best privacy cost gap obtained by DP-F is 68.4528\%. Likewise, for Thinwall with a privacy budget of 1, the AUC score of DP-Global-Adapt-V2 is 0.7687, while the second best score is achieved by DP-Global, which is 0.5668. We can see that the accuracy in Tables ~\ref{tab_4}-\ref{tab_8} is very high even when the privacy cost gap is high and the AUC score is low. This is because the datasets are extremely unbalanced and perform well in the classes with more samples. 

\begin{table}[]
\centering
\caption{Performance and Fairness metrics for DP-SGD.}
\label{tab_4}
\begin{tabular}{@{}lllllllll@{}}
\toprule
   & \multicolumn{4}{l}{Unbalanced MNIST}             & \multicolumn{4}{l}{Thinwall}                        \\ \midrule
$\epsilon$  & ACC.  & AUC & Group accuracy & $\pi$     & ACC.    & AUC & Group accuracy & $\pi$      \\ \midrule
1 & 97.59 & 0.9998 & {[}98.9341, 86.6530{]} & 9.3717 & 94.6292 & 0.5227 & {[}100.0000, 04.5454{]}      & 77.8148 \\
3  & 98.42 & 0.9999 & {[}99.3217, 92.9158{]} & 3.4965 & 95.6522 & 0.6136 & {[}100.0000, 22.7273{]}     & 59.6329 \\
5  & 98.64 & 0.9999 & {[}99.4186, 94.1478{]} & 2.3614 & 95.9079 & 0.7005 & {[}099.1870, 40.9091{]} & 40.6381 \\
8  & 98.73 & 0.9999 & {[}99.3217, 94.6611{]} & 1.7512 & 96.1637 & 0.7232 & {[}099.1870, 45.4545{]}  & 36.0927 \\
10 & 98.80 & 0.9999 & {[}99.4186, 95.0719{]} & 1.4373 & 96.1637 & 0.7232 & {[}099.1870, 45.4545{]} & 36.0927 \\ \bottomrule
\end{tabular}
\end{table}

\begin{table}[]
\centering
\caption{Performance and Fairness metrics for DP-F.}
\label{tab_5}
\begin{tabular}{@{}lllllllll@{}}
\toprule
   & \multicolumn{4}{l}{Unbalanced MNIST}             & \multicolumn{4}{l}{Thinwall}                        \\ \midrule
$\epsilon$  & ACC.  & AUC & Group accuracy & $\pi$ & ACC.    & AUC    & Group accuracy         & $\pi$ \\ \midrule
1  & 97.12 & 0.9997 & {[}99.0310, 82.6489{]} & 13.4727 & 94.6292 & 0.5227 & {[}100.0000, 04.5454{]}      & 77.8148           \\
3  & 98.18 & 0.9999 & {[}99.1279, 91.2731{]} & 04.9454   & 95.9079 & 0.7005 & {[}099.1870, 40.9090{]} & 40.6382 \\
5  & 98.52 & 0.9999 & {[}99.4186, 93.5318{]} & 02.9774   & 96.1637& 0.7232 & {[}099.1870, 45.4545{]}  & 36.0927 \\
8  & 98.58 & 0.9999 & {[}99.3217, 93.5183{]} & 02.8940   & 96.1637 & 0.7232 & {[}099.1870, 45.4545{]}  & 36.0927 \\
10 & 98.67 & 0.9999 & {[}99.5155, 94.4532{]} & 02.1529   & 96.1637 & 0.7660 & {[}098.6450, 54.5454{]} & 26.4598   \\ \bottomrule         
\end{tabular}
\end{table}

\begin{table}[]
\centering
\caption{Performance and Fairness metrics for DP-Global.}
\label{tab_6}
\begin{tabular}{@{}lllllllll@{}}
\toprule
   & \multicolumn{4}{l}{Unbalanced MNIST}             & \multicolumn{4}{l}{Thinwall}                        \\ \midrule
$\epsilon$  & ACC.  & AUC & Group accuracy & $\pi$    & ACC.    & AUC & Group accuracy & $\pi$      \\
1  & 87.29 & 0.9939 & {[}95.6395, 00.0000{]}    & 92.7301 & 94.8849 & 0.5668 & {[}99.7290, 13.6364{]}  & 68.4528 \\
3  & 88.67 & 0.9977 & {[}98.4496, 00.0000{]}    & 95.5402 &  95.6522 & 0.6123 & {[}99.7290, 22.7273{]} & 59.3619 \\
5  & 88.99 & 0.9982 & {[}98.8372, 00.0000{]}    & 95.9278 & 95.3900 & 0.6350 & {[}99.7290, 27.2727{]} & 54.8160 \\
8  & 89.12 & 0.9986 & {[}99.1279, 00.0000{]}    & 96.2185 & 95.3964 & 0.6550 & {[}99.1870, 31.8182{]} & 49.7290  \\
10 & 96.92 & 0.9997 & {[}99.2248, 79.4661{]}    & 16.8493 & 95.3964 & 0.6550 & {[}99.1870, 31.8182{]} & 49.7290  \\ \bottomrule
\end{tabular}
\end{table}

\begin{table}[]
\centering
\caption{Performance and Fairness metrics for DP-Global-Adapt.}
\label{tab_7}
\begin{tabular}{@{}lllllllll@{}}
\toprule
   & \multicolumn{4}{l}{Unbalanced MNIST}             & \multicolumn{4}{l}{Thinwall}                        \\ \midrule
$\epsilon$  & ACC.    & AUC & Group accuracy & $\pi$      & ACC.    & AUC & Group accuracy & $\pi$      \\
1  & 88.0200 & 0.9996 & {[}97.2868, 00.0000{]} & 94.3734 & 94.6292 & 0.5227 & {[}100.0000, 04.5454{]}      & 77.8148 \\
3 & 96.9000 & 0.9997 & {[}98.8372, 80.5955{]} & 15.3323 & 95.6522 & 0.6350 & {[}099.7290, 27.2727{]} & 54.8160 \\
5  & 97.8600 & 0.9998 & {[}99.4186, 88.0904{]} & 08.4188  & 95.9079 & 0.7005 & {[}099.1870, 40.9091{]} & 40.6381 \\
8  & 98.2000 & 0.9999 & {[}99.0310, 90.7598{]} & 05.3618  & 95.9079 & 0.7005 & {[}099.1870, 40.9091{]} & 40.6381 \\
10 & 98.4400   & 0.9999 & {[}99.3217, 92.6078{]} & 03.8045  & 95.9079 & 0.7219& {[}098.9160, 45.4545{]} & 35.8217 \\ \bottomrule
\end{tabular}
\end{table}

\begin{table}[]
\centering
\caption{Performance and Fairness metrics for DP-Global-Adapt-V2-S.}
\label{tab_8}
\begin{tabular}{@{}lllllllll@{}}
\toprule
   & \multicolumn{4}{l}{Unbalanced MNIST}             & \multicolumn{4}{l}{Thinwall}                        \\ \midrule
$\epsilon$  & ACC.    & AUC & Group accuracy & $\pi$     & ACC.    & AUC & Group accuracy & $\pi$      \\
1  & 98.7700 & 0.9999 & {[}98.9341, 95.0719{]} & 0.9528 & 96.6752 & 0.7687 & {[}99.1870, 54.5454{]} & 27.0018 \\
3  & 98.9200 & 0.9999 & {[}99.2248, 96.0986{]} & 0.2168 & 97.1867 & 0.9210 & {[}97.8320, 86.3636{]} & 06.1714 \\
5  & 99.0100 & 0.9999 & {[}99.7073, 95.9959{]} & 0.8020 & 97.1867& 0.9210 & {[}97.8320, 86.3636{]} & 06.1714 \\
8  & 99.0200 & 1.0000 & {[}99.6124, 96.4066{]} & 0.2964 & 97.4425 & 0.9223 & {[}98.1030, 86.3636{]} & 05.9004 \\
10 & 99.0500 & 1.0000 & {[}99.7093, 96.5092{]} & 0.2907 & 97.4425 & 0.9223 & {[}98.1030 , 86.3636{]} & 05.9004 \\ \bottomrule
\end{tabular}
\end{table}

\begin{table}[]
\centering
\caption{Performance and Fairness metrics for Non-DP.}
\label{tab_9}
\begin{tabular}{@{}llll@{}} 
\toprule
Dataset          & Accuracy & Overall\_AUC & Group accuracy          \\ \midrule
MNIST            & 99.3800\%,  & N/A          & N/A                     \\
CIFAR-10         & 95.2900\%,    & N/A          & N/A                     \\
CIFAR-100        & 80.4200\%,    & N/A          & N/A                     \\
Unbalanced MNIST & 99.1100\%,  & 1.000        & {[}99.8986\%, 96.5092\%{]}  \\
Thinwall         & 98.4655\%,  & 0.9064       & {[}99.4580\%,  81.8182\%{]}  \\ \bottomrule
\end{tabular}
\end{table} 

\subsection{Comparision on different types of DP-SGD-Global-Adapt-V2.} 

In this section, we compare DP-SGD with different types of DP-SGD-Global-Adapt-V2. Tables~\ref{tab_10}-\ref{tab_12} illustrate the performance using MNIST, CIFAR10, and CIFAR100. We run the experiments for the privacy budgets of 1, 3, 5, 8, and 10. Tables~\ref{tab_10}-\ref{tab_12} use Linear, Time, and Step to denote the different types of noise multiplier decay schedulers incorporated into DP-SGD-Global-Adapt-V2. In most cases, DP-SGD-Global-Adapt-V2-L and DP-SGD-Global-Adapt-V2-T have lower performance compared to DP-SGD. Using Tables~\ref{tab_10}-\ref{tab_12}, with a privacy budget of 1, DP-SGD achieves accuracy of 98.02\%, 94.30\%, and 76.75\%, while the (linear, time) methods achieve an accuracy of (97. 81\%, 97. 86\%), (94. 25\%, 94. 28\%), and (76. 67\%, 76. 73\%) in MNIST, CIFAR10, and CIFAR100, respectively. In some cases, the linear has higher performance than time, and in other cases, time has higher performance or performs the same as linear. To illustrate, consider Table~\ref{tab_10}, When the privacy budget is 1 and 3, the time decay has higher performance with an accuracy of 97.86\% and 98.70\%, while the linear decay has an accuracy of 97.81\% and 97.86\%, respectively. In contrast, when the privacy budget is 5 and 8, the linear decay has an accuracy of 98.91\% and 99.14\%, while the time decay has an accuracy of 98.89\% and 99.12\%,  respectively. At the privacy budget of 10, both linear and time decays have an accuracy of 99.15\%. In some cases, the linear decay or time decay has better performance than DP-SGD. For example, the time decay obtained an accuracy of 79.89\% and DP-SGD obtained an accuracy of 79.87\% at the privacy budget of 10 on CIFAR100. 

Similarly, on MNIST, linear decay has an accuracy of 99.14\%, while DP-SGD has an accuracy of 99.12\% at the privacy budget of 8. Time decay decays the noise multiplier in a similar way to linear decay. Our main contribution is the proposal of the step decaying noise multiplier and its integration with DP-SGD-Global-Adapt-V2. Across all datasets, DP-SGD-Global-Adapt-V2-S consistently outperforms DP-SGD, linear, and time by a substantial margin. In the following section, we will elucidate the reasons behind the superior performance of the step decay variant. 

\begin{table}[]
\centering
\caption{Comparision of different types of DP-SGD-Global-Adapt-V2 using MNIST.}
\label{tab_10}
\begin{tabular}{@{}lllll@{}}
\toprule
$\epsilon$  & DP-SGD & Linear & Time & Step \\ \midrule
1  & 98.02\%,  & 97.81\%,                    & 97.86\%,                    & 98.98\%     \\
3  & 98.77\%,  & 98.67\%,                    & 98.70\%,                    & 99.21\%      \\
5  & 98.98\%,  & 98.91\%,                    & 98.89\%,                    & 99.23\%          \\
8  & 99.12\%,  & 99.14\%,                    & 99.12\%,                    & 99.25\%                    \\
10 & 99.16\%,  & 99.15\%,                    & 99.15\%,                    & 99.26\%                   \\ \bottomrule
\end{tabular}
\end{table}

\begin{table}[]
\centering
\caption{Comparision on different types of DP-SGD-Global-Adapt-V2 using CIFAR10.}
\label{tab_11}
\begin{tabular}{@{}lllll@{}}
\toprule
$\epsilon$  & DP-SGD & Linear & Time & Step \\ \midrule
1  & 94.30\%,  & 94.25\%,                    & 94.28\%,                    & 94.95\%                    \\
3  & 94.76\%,  & 94.72\%,                    & 94.73\%,                    & 95.11\%                    \\
5  & 94.93 \%, & 94.93\%,                    & 94.91\%,                    & 95.18\%                    \\
8  & 94.96\%,  & 94.95 \%,                   & 94.96\%,                    & 95.22\%                    \\
10 & 95.04\%,  & 95.03\%,                    & 95.04\%,                   & 95.24\%                   \\ \bottomrule
\end{tabular}
\end{table}

\begin{table}[]
\centering
\caption{Comparision on different types of DP-SGD-Global-Adapt-V2 using CIFAR100.}
\label{tab_12}
\begin{tabular}{@{}lllll@{}}
\toprule
$\epsilon$  & DP-SGD & Linear & Time & Step \\ \midrule
1  & 76.75\%,  & 76.67\%,                    & 76.73\%,                    & 79.83\%,                    \\
3  & 78.63\%,  & 78.64\%,                    & 78.62\%,                    & 80.06\%,                    \\
5  & 79.40\%,   & 79.40\%,                    & 79.34\%,                    & 80.21\%,                    \\
8  & 79.74\%,  & 79.74\%,                    & 79.74\%,                    & 80.27\%,                    \\
10 & 79.87\%,  & 79.86\%,                    & 79.89\%,                    & 80.30\%,                   \\ \bottomrule
\end{tabular}
\end{table}

\subsection{Analysis on noise multiplier decay} 

\begin{figure}
\centering
\begin{center}
\includegraphics[width=1.0\textwidth,height=0.4\textheight,draft=false]{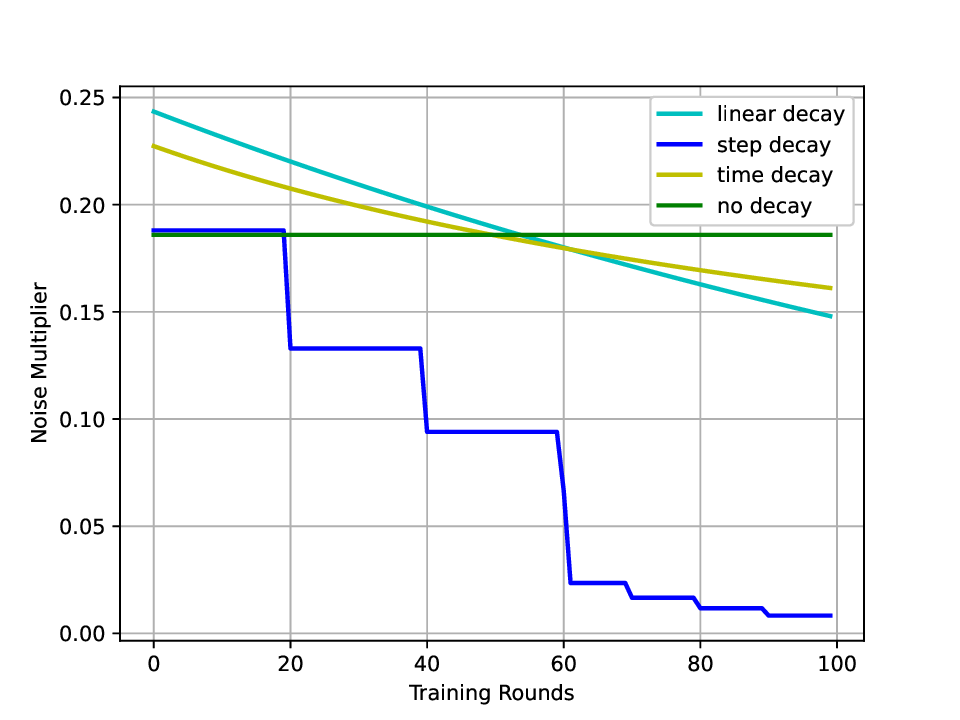}
\caption{Noise multiplier progression during training of DP-Global-Adapt-V2 for all the decay schedulers. We use the formulae shown in Table~\ref{tab_1} to compute the noise multiplier at every epoch (round). We use the drop rate of 0.99, 0.01, and 0.5 for linear, time, and step decay. For step decay, the step size (epoch drop rate) is 10.}
\label{noise_multiplier}
\end{center}
\end{figure}

\begin{figure}
\centering
\begin{center}
\includegraphics[width=1.0\textwidth,height=0.4\textheight,draft=false]{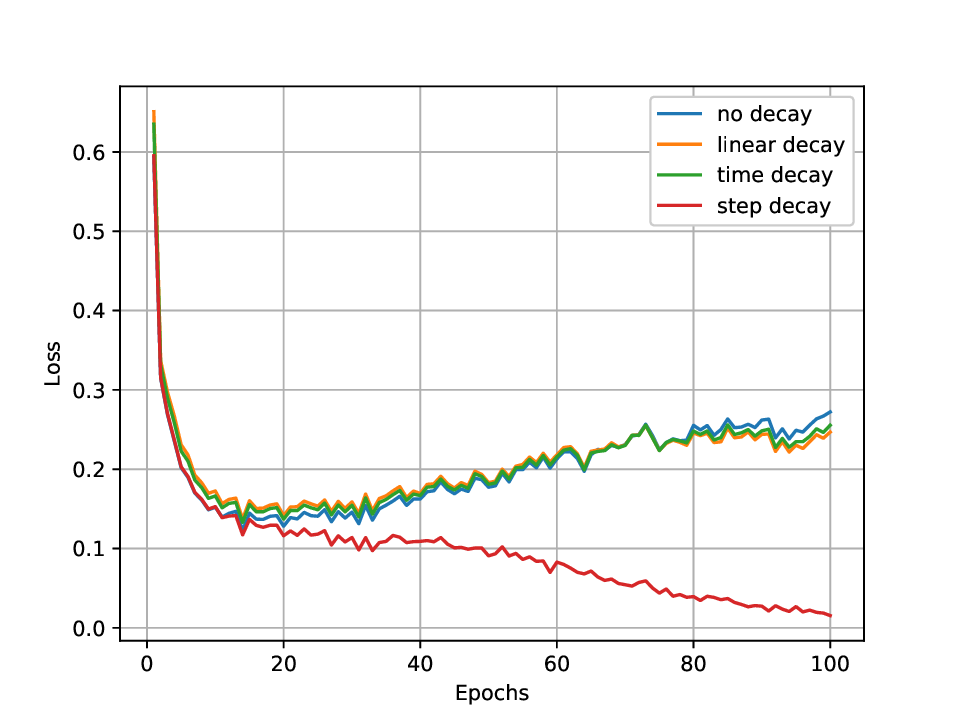}
\caption{Training loss over epochs for all the decay schedulers.  we use the MNIST dataset, AdamW optimizer, OCL LR scheduler, batch size of 64 and run the DP-SGD-Global-Adapt-V2 for 100 epochs of training and recorded the loss after every epoch.}
\label{loss}
\end{center}
\end{figure}

To analyze why the step decay variant performs better, we make Figures~\ref{noise_multiplier} and ~\ref{loss}. Fig.~\ref{noise_multiplier} describes the noise multiplier as the training progresses for the linear, time, step, and when no kind of decay is used. Fig.~\ref{loss} illustrates the loss as training progresses for the linear, time, step, and when no type of decay is used. We make the following interpretations based on Fig.~\ref{noise_multiplier}. Linear and time decay diffuse more noise into the model for almost half of the training rounds (50 training rounds) than the standard noise multiplier (no decay). This is the reason why, in most cases, the DP-SGD has higher performance compared to linear and time decay. Until 60 training rounds, the linear decay adds more noise to the model than the time decay, and after that, the time decay adds more noise than the linear decay. That is why, in some cases, time decay has better performance than linear decay, and in other cases, linear decay has better or equal performance to time decay. The step decay starts slightly higher than the standard noise multiplier and decreases rapidly at every step size (10 rounds), making the noise addition to the model at the end of training smaller. Rapid decrease in step decay and the ability to add less noise in most training rounds make step decay a better-performing algorithm. More importantly, we can select the best noise decay scheduler before training itself from Fig.~\ref{noise_multiplier} on the basis of noise addition to the model compared to the standard noise multiplier. 

We make the following observations from Fig.~\ref{loss}. The time, linear and no-decay algorithm's training losses decrease in initial training rounds and then start increasing again; this nature makes the model harder to converge. However, the step decay decreases as the training progresses and helps the model converge. So, to design better DP algorithms, the noise multiplier decay design is crucial to ensure the model convergence and give better utility and fairness. Since the step decay has performed better, in the following section we will discuss how to choose the hyperparameters for the step decay scheduler.

\subsection{How to choose the hyper-parameters of step decay noise multiplier.}

We used the MNIST dataset to execute DP-SGD-Global-Adapt-V2-S with varying step sizes and drop rates and the results are presented in Tables~\ref{tab_13} and ~\ref{tab_14}. Table~\ref{tab_13} illustrates how step size affects accuracy, while Table~\ref{tab_14} demonstrates the effect of drop rate on accuracy. From Table~\ref{tab_13}, we can observe that as the step size increases from 5 to 50, the noise multiplier and loss decrease, and the accuracy improves. Similarly, from Table~\ref{tab_14}, we can observe that as the drop rate increases from 0.1 to 0.9, the noise multiplier and loss decrease, and the accuracy improves. This is because DP algorithms are sensitive to the initial value of the noise multiplier. The lower the noise added to the model, the better the performance of the model. Therefore, to obtain a better DP model, we chose step decay with a larger step size and a higher drop rate.

\begin{table}[]
\centering
\caption{Analysis on step-size of step noise decay scheduler.}
\label{tab_13}
\begin{tabular}{@{}llll@{}}
\toprule
Step size & Noise multiplier & Accuracy & loss   \\ \midrule
5         & 8.6769           & 98.8000\%  & 0.0628 \\
10        & 0.1916           & 98.9800\%  & 0.0212 \\
20        & 0.0236           & 99.2500\%  & 0.0179 \\
25        & 0.0147           & 99.3000\%  & 0.0118 \\
50        & 0.0046           & 99.3200\%  & 0.0116 \\ \bottomrule
\end{tabular}
\end{table}

\begin{table}[]
\centering
\caption{Analysis of the drop rate of step noise decay scheduler.}
\label{tab_14}
\begin{tabular}{@{}llll@{}}
\toprule
Drop rate & Noise multiplier & Accuracy & loss   \\ \midrule
0.1       & 199.7241         & 41.9600\%  & 3.6502 \\
0.25      & 3.5423           & 91.1400\%  & 0.5526 \\
0.5       & 0.1916           & 98.9800\%  & 0.0496 \\
0.75      & 0.0425           & 99.0400\%  & 0.0212 \\
0.9       & 0.0246           & 99.1900\%  & 0.0186 \\ \bottomrule
\end{tabular}
\end{table}

\subsection{Analysis of model training hyper-parameters} \label{analysis hp}
This section examines the impact of training hyper-parameters, including the number of training rounds, batch size, optimizer, and learning rate scheduler, on the effectiveness of the DP model. We use the MNIST dataset and the privacy budget of 1 for this purpose.

\subsubsection{Analysis of the number of training rounds}

Table~\ref{tab_15} details the influence of different number of training iterations on the performance of DP-SGD-Global-Adapt-V2-S. The tests employ the Adamw optimizer, a batch size of 64, and a one-cycle learning rate policy. We explore five configurations of training rounds (10, 30, 50, 80, 100). Among these, 10 training rounds achieve the best results, with the 20 and 30 round sequences following in effectiveness. In contrast, the configuration with 100 training rounds is the fourth most effective, while the one with 80 rounds shows the poorest performance. In general, no definitive pattern emerges regarding how the number of training rounds affects the utility of DP-SGD-Global-Adapt-V2-S.

\begin{table}[]
\centering
\caption{Analysis of the number of training rounds.}
\label{tab_15}
\begin{tabular}{@{}lll@{}}
\toprule
\# training rounds & Accuracy & Loss   \\ \midrule
10                 & 99.22    & 0.0284 \\
30                 & 99.18    & 0.0216 \\
50                 & 99.07    & 0.0341 \\
80                 & 98.48    & 0.1100 \\
100                & 99.00    & 0.0168  \\ \bottomrule
\end{tabular}
\end{table}

\subsubsection{Analysis of batch size} 

Table~\ref{tab_16} presents how the varying batch sizes affect the performance of DP-SGD-Global-Adapt-V2-S. The AdamW optimizer is used, and we conduct 100 training rounds with a one-cycle learning rate scheduler. We tested batch sizes of 16, 32, 64, 128, and 256, assessing the performance of DP-Global-Adapt-V2-S. The findings do not reveal a clear trend with respect to how batch size impacts performance. A batch size of 64 proved to be most effective, followed by 16, 128, 32, and 256.

\begin{table}[]
\centering
\caption{Analysis of the batch size.}
\label{tab_16}
\begin{tabular}{@{}lll@{}}
\toprule
Batch size & Accuracy & Loss   \\ \midrule
16         & 98.02    & 0.3035 \\
32         & 97.93    & 0.1311 \\
64         & 99.00    & 0.0168 \\
128        & 97.97    & 0.1426 \\
256        & 95.53    & 0.3369 \\ \bottomrule
\end{tabular}
\end{table}

\subsubsection{Analysis of optimizer} 
Table~\ref{tab_17} illustrates how optimizers such as AdamW, Adam, SGD, and Rmsprop affect the performance of DP-SGD-Global-Adapt-V2-S. Of these optimizers, SGD exhibits notably poorer performance compared to the rest. AdamW achieves the best results, followed by RmsProp and Adam. 

\begin{table}[]
\centering
\caption{Analysis of the optimizer.}
\label{tab_17}
\begin{tabular}{@{}lll@{}} 
\toprule
Optimizer & Accuracy & Loss   \\ \midrule
AdamW     & 99.00    & 0.0168 \\
Adam      & 97.93    & 0.1889 \\
SGD       & 88.00    & 0.4308 \\
RmsProp   & 97.95    & 0.1361 \\ \bottomrule
\end{tabular}
\end{table}

\subsubsection{Analysis of the learning rate schedulers}

Table~\ref{tab_18} illustrates the effects of ten distinct learning rate schedulers (LR scheduler), including the one cycle learning rate (OCL), step (St), multi-step (Mst), constant (Con), linear (Li), exponential (Exp), cosine annealing (Cos), cosine annealing with warm restarts (CosWR), cyclic (Cyc), and reduce on plateau (ROP), on the efficacy of DP-SGD-Global-Adapt-V2-S. Among these, OCL demonstrates the highest performance, succeeded by Cos and CosWR, followed by Li, Con, Cyc, ROP, Mst, St, and finally Exp. It is important to highlight that the Exp LR scheduler exhibits notably poorer performance in comparison to the others.

\begin{table}\centering
\caption{Analysis of the learning rate schedulers.}
\label{tab_18}
\begin{tabular}{@{}lll@{}} \\ 
\toprule
LR scheduler & Accuracy & Loss   \\ \midrule
OCL          & 99.00    & 0.0168 \\
St           & 96.27    & 0.1954 \\
Mst          & 97.75    & 0.1326 \\
Con          & 98.17    & 0.1017 \\
Li           & 98.25    & 0.0970 \\
Exp          & 89.76    & 0.3925 \\
Cos          & 98.34    & 0.0911 \\
CosWR        & 98.34    & 0.0926 \\
Cyc          & 97.63    & 0.1490 \\
ROP          & 97.47    & 0.1333 \\ \bottomrule
\end{tabular}
\end{table}

\section{Convergence Analysis}
\begin{figure}
\centering
\begin{center}
\includegraphics[width=1.0\textwidth,height=0.4\textheight,draft=false]{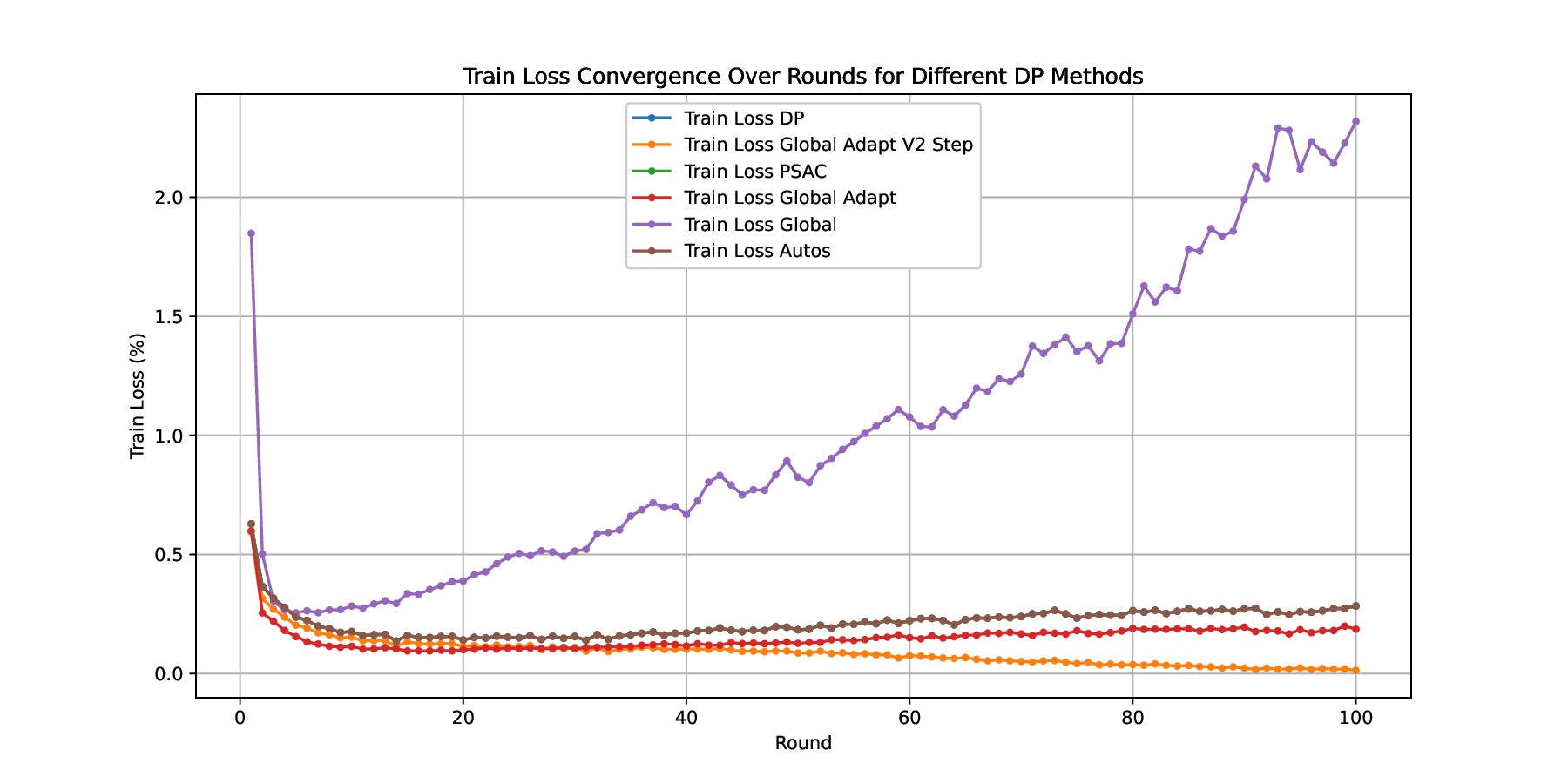}
\caption{Convergence analysis is performed on the MNIST dataset using train loss for DP-SGD, DP-PSAC, DP-Auto-s, DP-Global, DP-Global-Adapt, and DP-Global-Adapt-V2-S. Each algorithm undergoes 100 training epochs with a privacy budget of 1, utilizing the AdamW optimizer with a batch size of 64, alongside the OCL LR scheduler.}
\label{train_loss}
\end{center}
\end{figure}

\begin{figure}
\centering
\begin{center}
\includegraphics[width=1.0\textwidth,height=0.8\textheight,draft=false]{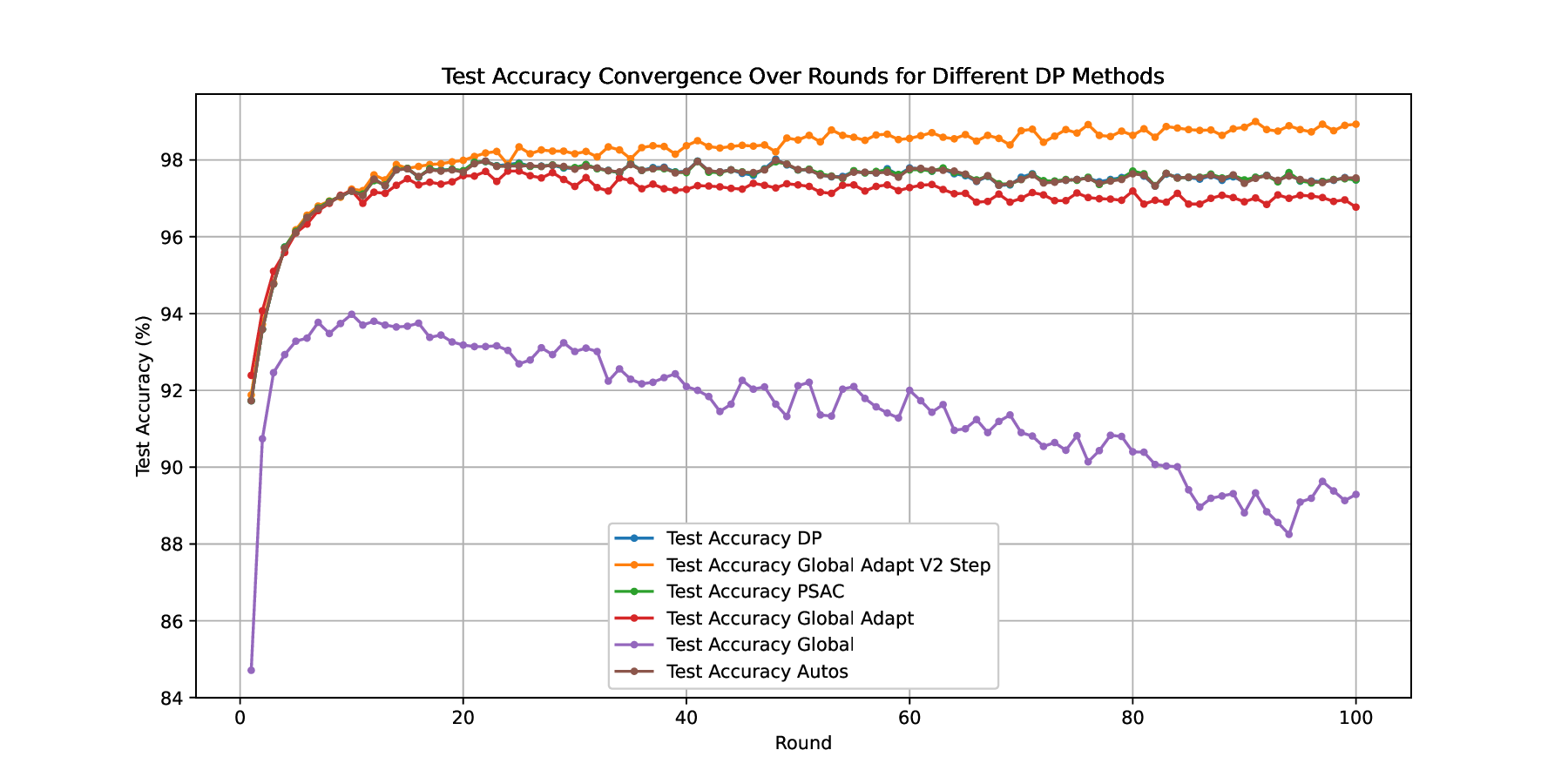}
\caption{Convergence analysis is performed on the MNIST dataset using test accuracy for DP-SGD, DP-PSAC, DP-Auto-s, DP-Global, DP-Global-Adapt, and DP-Global-Adapt-V2-S. Each algorithm undergoes 100 training epochs with a privacy budget of 1, utilizing the AdamW optimizer with a batch size of 64, alongside the OCL LR scheduler.}
\label{test_accuracy}
\end{center}
\end{figure}

This section presents the convergence behavior of the various DP methods examined in our experimental analysis within both the training and the test environments. Figures~\ref{train_loss}, ~\ref{test_accuracy} illustrate the convergence patterns of the DP algorithms trained to improve the accuracy of the model. Figure~\ref{train_loss} reveals that as training progresses, the train loss for DP-Global begins to diverge. In contrast, the other DP algorithms demonstrate convergence, with the proposed DP-Global-Adapt-V2-S achieving convergence at a lower loss level. Likewise, as illustrated in figure~\ref{test_accuracy}, the DP-Global-Adapt-V2-S algorithm we propose reaches convergence with greater accuracy, while the test accuracy for DP-Global decreases throughout the training. The other DP algorithms show similar performance.

\begin{figure}
\centering
\begin{center}
\includegraphics[width=1.0\textwidth,height=0.4\textheight,draft=false]{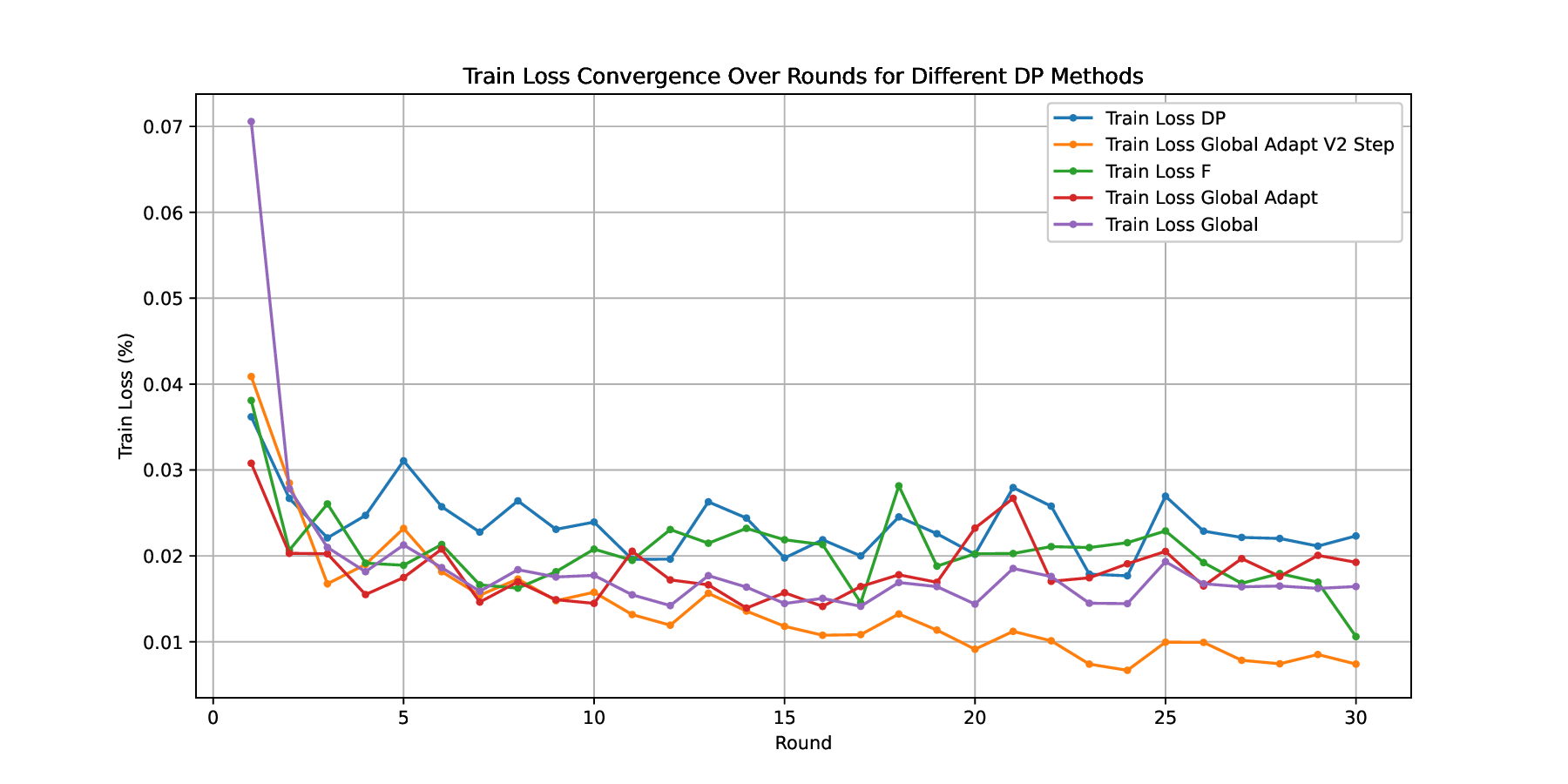}
\caption{Convergence analysis is performed on the Thinwall dataset using train loss for DP-SGD, DP-F, DP-Global, DP-Global-Adapt, and DP-Global-Adapt-V2-S. Each algorithm undergoes 30 training epochs with a privacy budget of 1, utilizing the AdamW optimizer with a batch size of 64, alongside the OCL LR scheduler.}
\label{train_loss_fair}
\end{center}
\end{figure}

\begin{figure}
\centering
\begin{center}
\includegraphics[width=1.0\textwidth,height=0.4\textheight,draft=false]{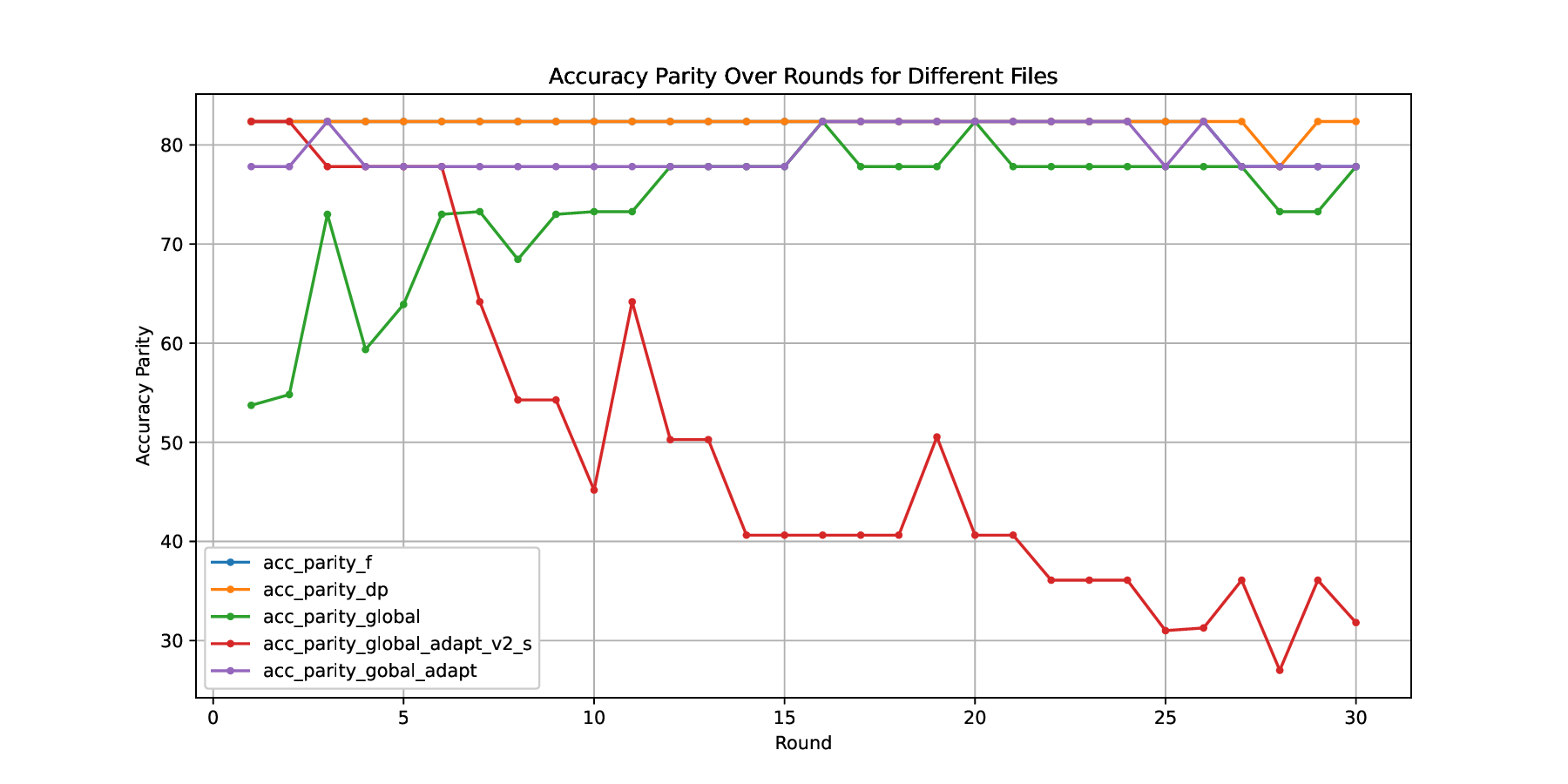}
\caption{Convergence analysis is performed on the Thinwall dataset using accuracy parity for DP-SGD, DP-F, DP-Global, DP-Global-Adapt, and DP-Global-Adapt-V2-S. Each algorithm undergoes 30 training epochs with a privacy budget of 1, utilizing the AdamW optimizer with a batch size of 64, alongside the OCL LR scheduler.}
\label{test_accuracy_parity}
\end{center}
\end{figure}

Figures~\ref{train_loss_fair}, \ref{test_accuracy_parity} illustrate the convergence performance of DP algorithms aimed at improving fairness in both the training and testing contexts. In Figure~\ref{train_loss_fair}, it is evident that the training loss for the suggested DP-Global-Adapt-V2-S is lower, while DP-SGD exhibits the highest loss. It is worth mentioning that the loss appears to fluctuate as its scale is in the order of $10^{-2}$. As depicted in figure~\ref{test_accuracy_parity}, the DP-Global-Adapt-V2-S exhibits a reduced accuracy parity, while DP-SGD demonstrates an increased accuracy parity. In general, in all cases, DP-Global-Adapt-V2-S shows better convergence and explains the reason behind its performance.

\section{Initial noise multiplier computation for a given privacy budget} 

\begin{figure}
\centering
\begin{center}
\includegraphics[width=1.0\textwidth,height=0.4\textheight,draft=false]{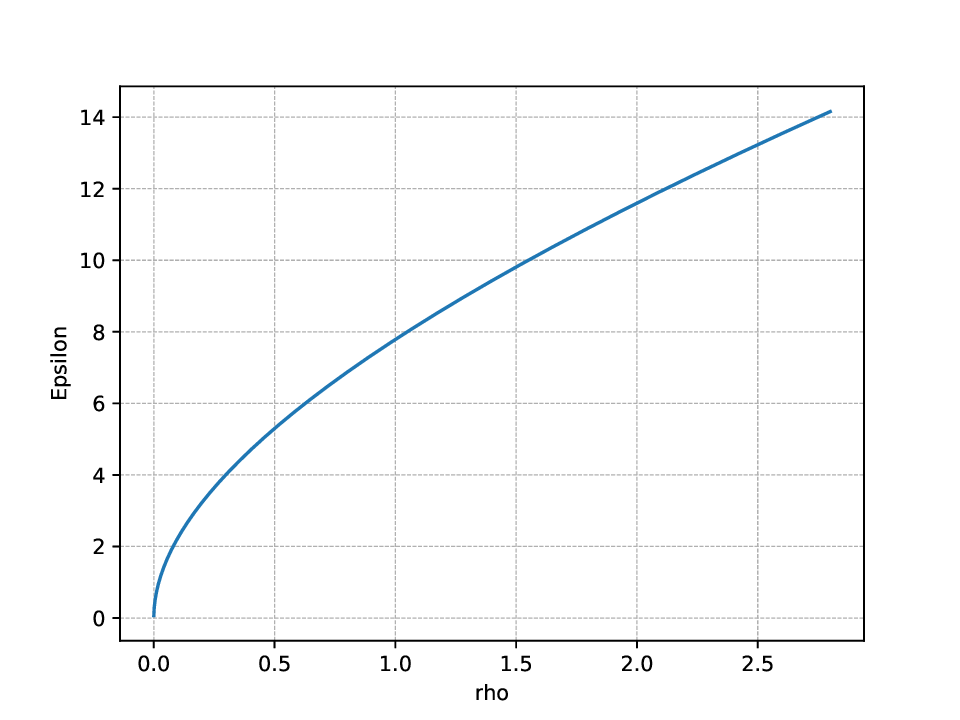}
\caption{Rho vs. epsilon for MNIST, CIFAR10, CIFAR100, and unbalanced MNIST. We use $\epsilon = (\rho_{total} +2\sqrt{\rho_{total} ln(1/\delta)})$ and $\delta$ is set to $1e-5$.}
\label{privacy}
\end{center}
\end{figure}

\begin{figure}
\centering
\begin{center}
\includegraphics[width=1.0\textwidth,height=0.4\textheight,draft=false]{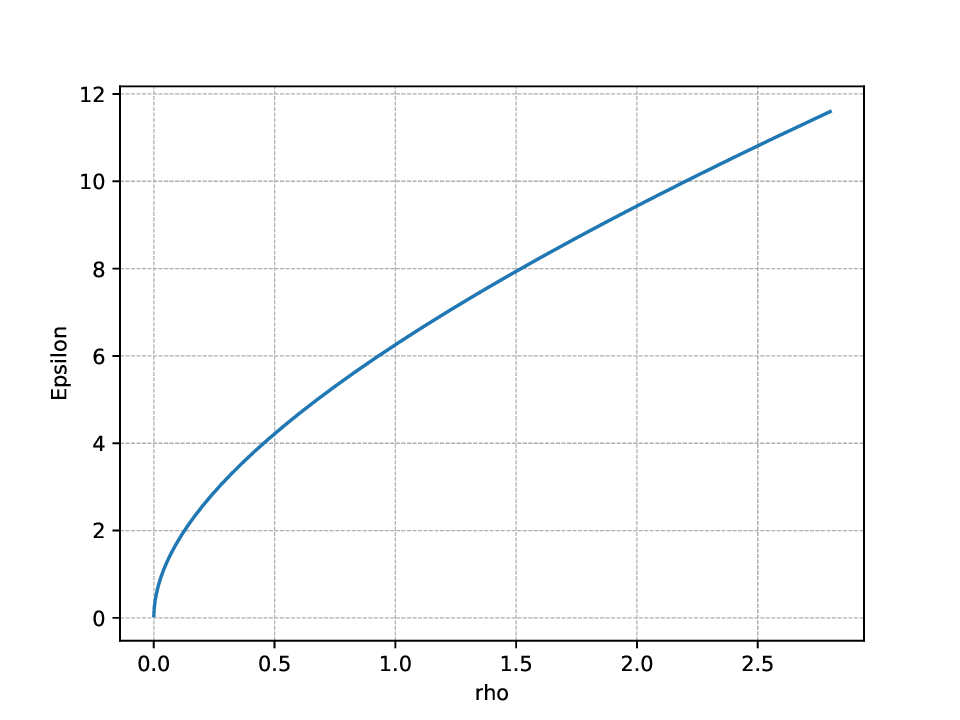}
\caption{Rho vs. epsilon for Thinwall. We use $\epsilon = (\rho_{total} +2\sqrt{\rho_{total} ln(1/\delta)})$ and $\delta$ is set to $1e-3$.}
\label{privacy2}
\end{center}
\end{figure}

Fig.~\ref{privacy} describes the relation between $\epsilon$ and $\rho_{total}$. We use Fig.~\ref{privacy} to obtain the initial noise multiplier for MNIST, CIFAR10, CIFAR100, and unbalanced MNIST. For a given privacy budget ($\epsilon$) using Fig.~\ref{privacy}, we compute $\rho_{total}$. Then, we use the expressions in Table~\ref{tab_9} to compute the initial noise multiplier for a respective noise multiplier decay scheduler. We follow a similar procedure to find the initial noise multiplier given the privacy budget ($\epsilon$) when using a Thinwall. The relation between $\epsilon$ and $\rho_{total}$ for Thinwall is shown in Fig.~\ref{privacy2}.

\section{Total privacy budget of DP-SGD-Global-Adapt-V2}
\label{8}
The notation $\log$ in all subsequent expressions refers to the natural logarithm. The sum of terms in a geometric sequence can be expressed as follows:
\begin{equation}
\label{eq:1}
s_n = \frac{a_1(r^n-1)}{r-1}, \quad r>1
\end{equation}
Here, $s_{n}$ represents the sum of the first $n$ terms of the geometric sequence, $r$ is the common ratio, $a_{1}$ denotes the first term in the geometric sequence, and $n$ represents the number of terms in the sequence.

The sum of the first $n$ natural numbers is expressed as follows:
\begin{equation}
\label{eq:2}
   \Sigma_{i=1}^{n}i = \frac{n(n+1)}{2}
\end{equation}

Based on Lemma ~\ref{lemma1} and Lemma ~\ref{lemma4}, the values of $\rho_{e}$ and $\omega_{e}$ are given by:
\begin{equation}
\label{x}
    \rho_{e} = 13(\frac{b}{n})^{2}(\frac{C^{2}}{2\sigma_{e}^2})
\end{equation}

\begin{equation}
\label{y}
    \omega_{e} = \frac{log(n/b)\sigma_{e}^2}{2C^{2}}
\end{equation}

Where $(b/n)$ represents the number of entries in a single batch of training examples and $e$ ranges from 0 to $E-1$.

Now, we can compute $\rho_{total}$ and $\omega_{total}$ using Lemma ~\ref{lemma2} as follows:

\begin{equation}
\label{eq : rho-total}
    \rho_{total} = 13(\frac{b}{n})^{2}(\frac{C^{2}}{2})(\frac{1}{\sigma_0^2} + \frac{1}{\sigma_1^2} + ... + \frac{1}{\sigma_e^2}+ ... + \frac{1}{\sigma_{E-1}^2})
\end{equation}

\begin{equation}
\label{eq4}
    \omega_{total} = \frac{log(n/b)}{2C^{2}}min(\sigma_{0}^{2}, \sigma_{1}^{2}, ...,\sigma_{e}{^2}, ..,\sigma_{E-1}^{2})
\end{equation}

\subsection{Derivation for DP-SGD-Global-Adapt-V2}

When the noise addition to the model is the same throughout the training, that is no noise multiplier decay is used, then equations ~\ref{x} and ~\ref{y} become:
\begin{equation}
\rho_{total} = 13(\frac{b}{n})^{2}(\frac{C^{2} E}{2\sigma^2})
\end{equation}

\begin{equation}
    \omega_{total} = \frac{log(n/b)\sigma^2}{2C^{2}}
\end{equation}

\subsection{Derivation for DP-SGD-Global-Adapt-V2-S}
To derive the final expression for the step decay, we simplify the step decay from Equation~\ref{st_1} to Equation~\ref{st_2} under the assumption that $E$ is divisible by $K$ and $P = E/K$. 
\begin{equation}
\label{st_1}
   \sigma_e^2 = \sigma_0^2 R^{\lfloor e/K \rfloor}, R\in(0,1)
\end{equation}

To illustrate the transformation of Equation~\ref{st_1}, consider $E = 100$, $K = 10$, and $P = \frac{E}{K} = \frac{100}{10} = 10$. Now, using Equation~\ref{st_1}, we can express $\sigma_e^2$ for $e$ which varies from 0 to $E-1 = 99$ as follows:

\begin{equation}
  \label{sigma_sum_1}
  \begin{array}{l}
   \sigma_0^2 = \sigma_0^2R^{\lfloor 0/10 \rfloor}=\sigma_0^2, \sigma_1^2 =\sigma_0^2 R^{\lfloor 1/10 \rfloor}=\sigma_0^2, ...,\sigma_{10}^2 = \sigma_0^2R^{\lfloor 10/10 \rfloor}=\sigma_0^2R, \\
   \sigma_{11}^2 =\sigma_0^2 R^{\lfloor 11/10 \rfloor} = \sigma_0^2R,..., \sigma_{98}^2 = \sigma_0^2R^{\lfloor 98/10 \rfloor} = \sigma_0^2R^9, \\
   \sigma_{99}^2 =\sigma_0^2 R^{\lfloor 99/10 \rfloor} = \sigma_0^2R^9
    \end{array}
\end{equation}

Now, the sum of the inverses of the noise multiplier at all epochs is equal to the following: 

\begin{equation}
\label{sigma_sum_2}
   \Sigma_{e=0}^{99}\frac{1}{\sigma_e^2} = \frac{10}{\sigma_0^2} + \frac{10}{R\sigma_0^2} + ... + \frac{10}{R^9\sigma_0^2}
\end{equation}

Equation~\ref{sigma_sum_2} can be generalized as follows:

\begin{equation}
\label{sigma_sum_3}
   \Sigma_{e=0}^{E-1}\frac{1}{\sigma_e^2} = \frac{K}{\sigma_0^2} + \frac{K}{R\sigma_0^2} + ... + \frac{K}{R^{p}\sigma_0^2}+ ... + \frac{K}{R^{P-1}\sigma_0^2}
\end{equation}

\begin{equation}
   \Sigma_{e=0}^{E-1}\frac{1}{\sigma_e^2} = \Sigma_{p=0}^{P-1}\frac{K}{R^{p}\sigma_0^2}
\end{equation}

To simplify the formula, let us define the following:

\begin{equation}
\label{sigma_sum_4}
   \Sigma_{e=0}^{E-1}\frac{1}{\sigma_e^2} = \Sigma_{p=0}^{P-1}\frac{1}{\sigma_p^2}
\end{equation}

Then, $\sigma_p^2$ can be expressed as follows:
\begin{equation}
\label{st_2}
    \sigma_p^2 = \frac{R^p \sigma_0^2}{K}
\end{equation}

Equation~\ref{eq : rho-total} can be generalized as follows: 

\begin{equation}
\label{eq : rho-total3}
    \rho_{total} = 13(\frac{b}{n})^{2}(\frac{C^{2}}{2})(\Sigma_{p=0}^{P-1}\frac{1}{\sigma_p^2})
\end{equation}

Now, let us substitute Equation~\ref{st_2} into Equation~\ref{eq : rho-total3}:

\begin{equation}
\label{eq : rho-total4}
    \rho_{total} = 13(\frac{b}{n})^{2}(\frac{C^{2}K}{2\sigma_0^2})(\Sigma_{p=0}^{P-1}\frac{1}{R^p})
\end{equation}

In expanding Equation~\ref{eq : rho-total4}, it becomes as follows:

\begin{equation}
\label{step1}
   \rho_{total} = 13(\frac{b}{n})^{2}(\frac{C^{2}K}{2\sigma_0^2})(1 + \frac{1}{R} + \frac{1}{R^{2}} + ... +\frac{1}{R^{p}}+ ... + \frac{1}{R^{P-1}})
\end{equation}

Using the formula for the sum of terms in a geometric sequence, Equation~\ref{eq:1}, we can summarize Equation~\ref{step1} as follows:

\begin{equation}
\label{step2}
    \rho_{total} = 13(\frac{b}{n})^{2}(\frac{C^{2}K}{2\sigma_0^2})(\frac{1.(\frac{1}{R})^{P}-1}{\frac{1}{R} - 1})
\end{equation}

After further simplifying Equation~\ref{step2}:

\begin{equation}
\label{step3}
       \rho_{total} = 13(\frac{b}{n})^{2}(\frac{C^{2}D}{2\sigma_0^2})(\frac{1-R^{P}}{R^{P-1}-{R^{P}}})
\end{equation}

Now, let us substitute the revised step noise multiplier decay (Equation~\ref{st_2}) into Equation~\ref{step3}:

\begin{equation}
    \omega_{total} = \frac{log(n/b)}{2C^{2}}min(\frac{\sigma_{0}^{2}}{K}, \frac{\sigma_{0}^{2}}{K}, ..., \frac{R^{p}\sigma_{0}^{2}}{K}, ..., \frac{R^{P-1}\sigma_{0}^{2}}{K})
\end{equation}

\begin{equation}
    \omega_{total} = \frac{log(n/b)\sigma_{0}^{2}}{2C^{2}K}min(1, R, ..., R^{p}, ..., R^{P-1})
\end{equation}

Since $R < 1$, we have:

\begin{equation}
    \omega_{total} = \frac{log(n/b)\sigma_{0}^{2}}{2C^{2}K}(R^{P-1})
\end{equation}

\subsection{Derivation for DP-SGD-Global-Adapt-V2-L}
According to the linear noise multiplier decay mechanism~\cite{zhang2021adaptive}: 

\begin{equation}
\label{linear}
    \sigma_{e}^{2} = R\sigma_{e-1}^2, R \in (0,1)
\end{equation}                               

Now, let us substitute Equation~\ref{linear} into Equation~\ref{eq : rho-total}:

\begin{equation}
   \rho_{total} = 13(\frac{b}{n})^{2}(\frac{C^{2}}{2})(\frac{1}{\sigma_{0}^{2}} + \frac{1}{R\sigma_0^2} + \frac{1}{R^{2}\sigma_0^2} + ... + \frac{1}{R^{e}\sigma_0^2} +...+ \frac{1}{R^{E-1}\sigma_0^2})
\end{equation}

\begin{equation}
\label{rho-gm}
   \rho_{total} = 13(\frac{b}{n})^{2}(\frac{C^{2}}{2\sigma_0^2})(1 + \frac{1}{R} + \frac{1}{R^{2}} + ... + \frac{1}{R^{e}} +... + \frac{1}{R^{E-1}})
\end{equation}

Equation~\ref{rho-gm} can be summarized as follows:

\begin{equation}
\label{eq:linear1}
   \rho_{total} = 13(\frac{b}{n})^{2}(\frac{C^{2}}{2\sigma_0^2})(\Sigma_{e=0}^{E-1}\frac{1}{R^{e}})
\end{equation}

Using the sum of the terms of the geometric sequence formula~\ref{eq:1}, the equation~\ref{eq:linear1} can be summarized as follows:

\begin{equation}
\label{eq:8}
    \rho_{total} = 13(\frac{b}{n})^{2}(\frac{C^{2}}{2\sigma_0^2})(\frac{1.(\frac{1}{R})^{E}-1}{\frac{1}{R} - 1})
\end{equation}

After simplifying the equation~\ref{eq:8} further:

\begin{equation}
\label{eq:9}
       \rho_{total} = 13(\frac{b}{n})^{2}(\frac{C^{2}}{2\sigma_0^2})(\frac{1-R^{E}}{R^{E-1}-{R^{E}}})
\end{equation}

Now, substitute the linear noise multiplier decay~\ref{linear} into the equation~\ref{eq4}

\begin{equation}
    \omega_{total} = \frac{log(n/b)}{2C^{2}}min(\sigma_{0}^{2}, R\sigma_{0}^{2}, ..., R^{e}\sigma_{0}^{2}, ..., R^{E-1}\sigma_{0}^{2})
\end{equation}

\begin{equation}
   \label{omega_1}
    \omega_{total} = \frac{log(n/b)\sigma_{0}^{2}}{2C^{2}}min(1, R, ..., R^{e}, ..., R^{E-1})
\end{equation}

Since $R \in (0, 1)$, equation~\ref{omega_1} becomes as follows:

\begin{equation}
    \omega_{total} = \frac{log(n/b)\sigma_{0}^{2}}{2C^{2}}(R^{E-1})
\end{equation}

\subsection{Derivation for DP-SGD-Global-Adapt-V2-T}

The time noise multiplier decay mechanism is expressed as follows: 
\begin{equation}
\label{time}
    \sigma_{e}^{2} = \frac{\sigma_{0}^2}{1+Re}, R = \frac{\sigma_0}{T}, R \in (0,1)
\end{equation}

Substituting equation~\ref{time} in equation~\ref{eq : rho-total}. Equation~\ref{eq : rho-total} looks like follows:

\begin{equation}
\label{eq13}
   \rho_{total} = 13(\frac{b}{n})^{2}\frac{C^{2}}{2}(\frac{1}{\sigma_{0}^2} + \frac{1+R}{\sigma_{0}^2} + \frac{1+2R}{\sigma_{0}^2} + ... +\frac{1+Re}{\sigma_{0}^2}+   
   ... + \frac{1+R(E-1)}{\sigma_{0}^2})
\end{equation}

\begin{equation}
\label{eq13-1}
   \rho_{total} = 13(\frac{b}{n})^{2}(\frac{C^{2}}{2\sigma_0^2})[1 + (1 + R) + (1 + 2R) + ... +(1 + eR)+... + (1 + (E-1)R)]
\end{equation}

The equation~\ref{eq13-1} can be summarized as follows:
\begin{equation}
\label{eq13-2}
   \rho_{total} = 13(\frac{b}{n}^{2}(\frac{C^{2}}{2\sigma_0^2})[\Sigma_{e=0}^{E-1}(1 + Re)]
\end{equation}

In simplifying, equation~\ref{eq13-2} using equation~\ref{eq:2} becomes as follows:

\begin{equation}
   \rho_{total} = 13(\frac{b}{n})^{2}(\frac{C^{2}}{2\sigma_0^2})[E + \frac{R(E)(E-1)}{2}]
\end{equation}

\begin{equation}
\label{eq14}
   \rho_{total} = 13(\frac{b}{n})^{2}(\frac{C^{2}}{4\sigma_0^2})[2E + R(E)(E-1)]
\end{equation}

Now, substitute the decay of the time noise multiplier~\ref{time} into the equation~\ref{eq4}

\begin{equation}
    \omega_{total} = \frac{log(n/b)}{2C^{2}}min(\sigma_{0}^{2}, \frac{\sigma_{0}^{2}}{1+R}, ..., \frac{\sigma_{0}^{2}}{1+Re}, ..., \frac{\sigma_{0}^{2}}{1+R(E-1)})
\end{equation}

\begin{equation}
\label{eq15}
    \omega_{total} = \frac{log(n/b)\sigma_{0}^{2}}{2C^{2}}min(1, \frac{1}{1+R}, ..., \frac{1}{1+Re}, ..., \frac{1}{1+R(E-1)})
\end{equation}

In further simplification, Equation~\ref{eq15} becomes:

\begin{equation}                                                                                                                    
\label{eq16}
    \omega_{total} = \frac{log(n/b)\sigma_{0}^{2}}{2C^{2}(1+R(E-1))}
\end{equation}

\section{The Importance of Differential Privacy (DP) in Additive Manufacturing (AM)}
\label{B}

Differential Privacy (DP) protects sensitive information by adding noise while allowing meaningful analysis~\cite{hassan2019differential, jiang2021differential}. Its adaptability to various types of data and applications makes it useful for protecting confidential information not only in additive manufacturing (AM), but also in other domains~\cite{gartner2022local, jain2018differential}. Due to the increasing accessibility and prevalence of 3D printing technology, the need for DP becomes increasingly evident for some key reasons. Firstly, 3D printing allows for easy replication of physical objects based on digital designs or process data~\cite{balletti20173d, jandyal20223d}. However, without adequate data privacy measures, these digital designs and processes can be vulnerable to theft or unauthorized replication, posing a significant risk to intellectual property rights~\cite{fullington2023design}. 
In this regard, DP techniques such as DP-SGD~\cite{abadi2016deep} can obscure subtle details within designs, prevent reverse engineering attempts, and protect proprietary information from AM processes~\cite{owusu2023msdp}. This ensures that sensitive designs remain undisclosed to unauthorized parties, preserving competitive advantages and improving trust among stakeholders~\cite{dwork2014algorithmic}. 
Specifically, in the AM dataset, anonymity can be achieved by eliminating specific design details, such as the trajectory of the printing path, and effectively protecting proprietary information while maintaining data integrity and confidentiality. 

The major benefits of implementing DP in AM process monitoring can be listed as follows: (i) Enhanced Collaboration: By providing confidence in the protection of sensitive information, the adoption of DP frameworks encourages collaboration among industries, researchers, and designers. Secure data sharing facilitates knowledge transfer and accelerates innovation within intelligent AM environments~\cite{narayanan2019robust}. ii) Regulatory compliance: Global data privacy issues are causing regulatory agencies to closely examine data handling procedures across a range of businesses, including AM. Implementing DP measures not only aligns with emerging regulatory requirements but also demonstrates a commitment to ethical data management and corporate responsibility~\cite{gil2019understanding}. Ultimately, DP is essential to ensure the responsible and secure use of data in AM. By prioritizing privacy-preserving techniques, stakeholders can protect the integrity of sensitive information, promote collaboration, and navigate evolving regulatory environments with confidence. In addition, as AM continues to redefine production paradigms, DP emerges as a vital tool for protecting innovation and driving sustainable growth in this digital era.

\section{Limitations and Future Work} 
The DP-SGD-Global-Adapt-L and DP-SGD-Global-Adapt-T perform worse than DP-SGD in most cases because more noise is added to the model in more than half of the training stages than DP-SGD. One potential area for improvement involves the noise multiplier decay function. A more flexible approach would be to design a decay function that allows for customization of the noise multiplier reduction at specific iterations rather than relying on predefined steps like those used in linear and time-based methods (every epoch) or step-based methods (after 10 epochs). Designing the noise multiplier dynamically according to the characteristics of the data set and the model architecture could be a good direction to improve DP-SGD-Global-Adapt-V2. The privacy accountant tCDP can be replaced with an exact computation method, such as numerical methods~\cite{gopi2021numerical}, to obtain an exact computation of privacy. However, using it is not straightforward when the noise multiplier changes during training. Another interesting direction could be searching for the optimal model architecture using a differentially private neural architecture search. A different method to enhance fairness while preserving privacy is to use generative deep learning to create synthetic samples in equal proportions across all sample categories. This research introduces promising opportunities to integrate the suggested clipping mechanism and step-decaying noise multiplier into DP-based generative AI models, including GANs and diffusion models. This integration has the potential to significantly improve the privacy, utility, and fairness trade-offs in these approaches.

\section{Conclusion}

We show that existing DP algorithms perform either worse or equally well in most cases and perform better only in some cases. In addition, we explain the reasons for the poor performance of existing work, such as DP-Global and DP-Global-Adapt, by showing the trend of the upper clipping threshold. To improve the privacy, utility, and fairness trade-off, we designed the DP-SGD-Global-Adapt-V2-S, which uses a step decay noise multiplier and step decay based clipping threshold.  We conducted an extensive evaluation using the five different datasets, such as MNIST, CIFAR10, CIFAR100, unbalanced MNIST, and Thinwall. Specifically, DP-SGD-Global-Adapt-V2-S at the privacy budget ($\epsilon$) of 1, improves accuracy by 0.9795\%, 0.6786\%, and 4.0130\% in MNIST, CIFAR10, and CIFAR100, respectively, and reduces the privacy cost gap ($\pi$) by 89.8332\% and 60.5541\% in unbalanced MNIST and Thinwall, respectively. We develop a mathematical expression to compute the privacy budget of the proposed algorithm using tCDP. We also provided an analysis of why the step decay noise multiplier has higher performance and provided a recommendation on choosing its hyperparameters. We discussed about the results of the different noise multiplier scheduler of DP-Global-Adapt-V2. More importantly, we also demonstrate the convergence behavior for all DP algorithms and also analyze the results of model training hyperparameters.

\section{Declaration of generative AI and AI-assisted technologies}
During the preparation of this work, we used Grammarly and Writefull for paraphrasing and grammar correction. After using these services, we reviewed and edited the content as needed and took full responsibility for the content of the publication.

\section{Acknowledgement}

The funding parts do not play any role in the research. There are four funds to be acknowledged. This work is supported in part by the US NSF under grants OIA-1946231, CNS-2117785, OIA-2229752, and CNS-2231682.

\bibliographystyle{elsarticle-num}

\begin{thebibliography}{10}
\expandafter\ifx\csname url\endcsname\relax
  \def\url#1{\texttt{#1}}\fi
\expandafter\ifx\csname urlprefix\endcsname\relax\def\urlprefix{URL }\fi
\expandafter\ifx\csname href\endcsname\relax
  \def\href#1#2{#2} \def\path#1{#1}\fi

\bibitem{sarker2021deep}
I.~H. Sarker, Deep learning: a comprehensive overview on techniques, taxonomy, applications and research directions, SN Computer Science 2~(6) (2021) 420.

\bibitem{ardila2019end}
D.~Ardila, A.~P. Kiraly, S.~Bharadwaj, B.~Choi, J.~J. Reicher, L.~Peng, D.~Tse, M.~Etemadi, W.~Ye, G.~Corrado, et~al., End-to-end lung cancer screening with three-dimensional deep learning on low-dose chest computed tomography, Nature medicine 25~(6) (2019) 954--961.

\bibitem{huang2020deep}
J.~Huang, J.~Chai, S.~Cho, Deep learning in finance and banking: A literature review and classification, Frontiers of Business Research in China 14~(1) (2020) 1--24.

\bibitem{healthcaredive}
\url{https://www.healthcaredive.com/news/artificial-intelligence-healthcare-savings-harvard-mckinsey-report/641163/}, [accessed on 2-Feb-2023] (2023).

\bibitem{FinancialServices}
\url{https://www.insiderintelligence.com/insights/ai-in-finance}, [accessed on 14-Mar-2023] (2023).

\bibitem{hassani2020deep}
H.~Hassani, X.~Huang, E.~Silva, M.~Ghodsi, Deep learning and implementations in banking, Annals of Data Science 7 (2020) 433--446.

\bibitem{shokri2017membership}
R.~Shokri, M.~Stronati, C.~Song, V.~Shmatikov, Membership inference attacks against machine learning models, in: 2017 IEEE symposium on security and privacy (SP), IEEE, 2017, pp. 3--18.

\bibitem{hu2022membership}
H.~Hu, Z.~Salcic, L.~Sun, G.~Dobbie, P.~S. Yu, X.~Zhang, Membership inference attacks on machine learning: A survey, ACM Computing Surveys (CSUR) 54~(11s) (2022) 1--37.

\bibitem{truex2018towards}
S.~Truex, L.~Liu, M.~E. Gursoy, L.~Yu, W.~Wei, Towards demystifying membership inference attacks, arXiv preprint arXiv:1807.09173 (2018).

\bibitem{gong2016you}
N.~Z. Gong, B.~Liu, You are who you know and how you behave: Attribute inference attacks via users' social friends and behaviors., in: USENIX Security Symposium, 2016, pp. 979--995.

\bibitem{gong2018attribute}
N.~Z. Gong, B.~Liu, Attribute inference attacks in online social nerks, ACM Transactions on Privacy and Security (TOPS) 21~(1) (2018) 1--30.

\bibitem{zhao2021feasibility}
B.~Z.~H. Zhao, A.~Agrawal, C.~Coburn, H.~J. Asghar, R.~Bhaskar, M.~A. Kaafar, D.~Webb, P.~Dickinson, On the (in) feasibility of attribute inference attacks on machine learning models, in: 2021 IEEE European Symposium on Security and Privacy (EuroS\&P), IEEE, 2021, pp. 232--251.

\bibitem{fredrikson2015model}
M.~Fredrikson, S.~Jha, T.~Ristenpart, Model inversion attacks that exploit confidence information and basic countermeasures, in: Proceedings of the 22nd ACM SIGSAC conference on computer and communications security, 2015, pp. 1322--1333.

\bibitem{wu2016methodology}
X.~Wu, M.~Fredrikson, S.~Jha, J.~F. Naughton, A methodology for formalizing model-inversion attacks, in: 2016 IEEE 29th Computer Security Foundations Symposium (CSF), IEEE, 2016, pp. 355--370.

\bibitem{chen2020improved}
S.~Chen, R.~Jia, G.-J. Qi, Improved techniques for model inversion attacks (2020).

\bibitem{dwork2014dwork}
R.~Dwork, et~al., Dwork c., roth a, The algorithmic foundations of differential privacy, Foundations and Trends in Theoretical Computer Science 9~(3-4) (2014) 211--407.

\bibitem{dinur2003revealing}
I.~Dinur, K.~Nissim, Revealing information while preserving privacy, in: Proceedings of the twenty-second ACM SIGMOD-SIGACT-SIGART symposium on Principles of database systems, 2003, pp. 202--210.

\bibitem{near2021programming}
J.~P. Near, C.~Abuah, Programming differential privacy, URL: https://uvm (2021).

\bibitem{sweeney2015only}
L.~Sweeney, Only you, your doctor, and many others may know, Technology Science 2015092903~(9) (2015) 29.

\bibitem{dwork2011differential}
C.~Dwork, F.~McSherry, K.~Nissim, A.~Smith, Differential privacy—a primer for the perplexed,”, Joint UNECE/Eurostat work session on statistical data confidentiality 11 (2011).

\bibitem{dwork2006calibrating}
C.~Dwork, F.~McSherry, K.~Nissim, A.~Smith, Calibrating noise to sensitivity in private data analysis, in: Theory of Cryptography: Third Theory of Cryptography Conference, TCC 2006, New York, NY, USA, March 4-7, 2006. Proceedings 3, Springer, 2006, pp. 265--284.

\bibitem{farrand2020neither}
T.~Farrand, F.~Mireshghallah, S.~Singh, A.~Trask, Neither private nor fair: Impact of data imbalance on utility and fairness in differential privacy, in: Proceedings of the 2020 workshop on privacy-preserving machine learning in practice, 2020, pp. 15--19.

\bibitem{abadi2016deep}
M.~Abadi, A.~Chu, I.~Goodfellow, H.~B. McMahan, I.~Mironov, K.~Talwar, L.~Zhang, Deep learning with differential privacy, in: Proceedings of the 2016 ACM SIGSAC conference on computer and communications security, 2016, pp. 308--318.

\bibitem{chen2020understanding}
X.~Chen, S.~Z. Wu, M.~Hong, Understanding gradient clipping in private sgd: A geometric perspective, Advances in Neural Information Processing Systems 33 (2020) 13773--13782.

\bibitem{zhang2021adaptive}
X.~Zhang, J.~Ding, M.~Wu, S.~T. Wong, H.~Van~Nguyen, M.~Pan, Adaptive privacy preserving deep learning algorithms for medical data, in: Proceedings of the IEEE/CVF Winter Conference on Applications of Computer Vision, 2021, pp. 1169--1178.

\bibitem{bu2024automatic}
Z.~Bu, Y.-X. Wang, S.~Zha, G.~Karypis, Automatic clipping: Differentially private deep learning made easier and stronger, Advances in Neural Information Processing Systems 36 (2024).

\bibitem{yang2022normalized}
X.~Yang, H.~Zhang, W.~Chen, T.-Y. Liu, Normalized/clipped sgd with perturbation for differentially private non-convex optimization, arXiv preprint arXiv:2206.13033 (2022).

\bibitem{xia2023differentially}
T.~Xia, S.~Shen, S.~Yao, X.~Fu, K.~Xu, X.~Xu, X.~Fu, Differentially private learning with per-sample adaptive clipping, in: Proceedings of the AAAI Conference on Artificial Intelligence, Vol.~37, 2023, pp. 10444--10452.

\bibitem{esipova2022disparate}
M.~S. Esipova, A.~A. Ghomi, Y.~Luo, J.~C. Cresswell, Disparate impact in differential privacy from gradient misalignment, arXiv preprint arXiv:2206.07737 (2022).

\bibitem{bu2021convergence}
Z.~Bu, H.~Wang, Q.~Long, On the convergence and calibration of deep learning with differential privacy, arXiv preprint arXiv:2106.07830 (2021).

\bibitem{zhu2023informational}
P.~Zhu, C.~Miao, Z.~Wang, X.~Li, Informational cascade, regulatory focus and purchase intention in online flash shopping, Electronic Commerce Research and Applications 62 (2023) 101343.

\bibitem{zhu2024novel}
P.~Zhu, H.~Zhang, Y.~Shi, W.~Xie, M.~Pang, Y.~Shi, A novel discrete conformable fractional grey system model for forecasting carbon dioxide emissions, Environment, Development and Sustainability (2024) 1--29.

\bibitem{cai2022deep}
Y.~Cai, W.~Ke, E.~Cui, F.~Yu, A deep recommendation model of cross-grained sentiments of user reviews and ratings, Information Processing \& Management 59~(2) (2022) 102842.

\bibitem{xu2021removing}
D.~Xu, W.~Du, X.~Wu, Removing disparate impact on model accuracy in differentially private stochastic gradient descent, in: Proceedings of the 27th ACM SIGKDD Conference on Knowledge Discovery \& Data Mining, 2021, pp. 1924--1932.

\bibitem{dwork2006our}
C.~Dwork, K.~Kenthapadi, F.~McSherry, I.~Mironov, M.~Naor, Our data, ourselves: Privacy via distributed noise generation, in: Advances in Cryptology-EUROCRYPT 2006: 24th Annual International Conference on the Theory and Applications of Cryptographic Techniques, St. Petersburg, Russia, May 28-June 1, 2006. Proceedings 25, Springer, 2006, pp. 486--503.

\bibitem{dwork2009differential}
C.~Dwork, J.~Lei, Differential privacy and robust statistics, in: Proceedings of the forty-first annual ACM symposium on Theory of computing, 2009, pp. 371--380.

\bibitem{dwork2010boosting}
C.~Dwork, G.~N. Rothblum, S.~Vadhan, Boosting and differential privacy, in: 2010 IEEE 51st Annual Symposium on Foundations of Computer Science, IEEE, 2010, pp. 51--60.

\bibitem{mironov2017renyi}
I.~Mironov, R{\'e}nyi differential privacy, in: 2017 IEEE 30th computer security foundations symposium (CSF), IEEE, 2017, pp. 263--275.

\bibitem{dong2019gaussian}
J.~Dong, A.~Roth, W.~J. Su, Gaussian differential privacy, arXiv preprint arXiv:1905.02383 (2019).

\bibitem{bun2018composable}
M.~Bun, C.~Dwork, G.~N. Rothblum, T.~Steinke, Composable and versatile privacy via truncated cdp, in: Proceedings of the 50th Annual ACM SIGACT Symposium on Theory of Computing, 2018, pp. 74--86.

\bibitem{gopi2021numerical}
S.~Gopi, Y.~T. Lee, L.~Wutschitz, Numerical composition of differential privacy, Advances in Neural Information Processing Systems 34 (2021) 11631--11642.

\bibitem{dwork2008differential}
C.~Dwork, Differential privacy: A survey of results, in: Theory and Applications of Models of Computation: 5th International Conference, TAMC 2008, Xi’an, China, April 25-29, 2008. Proceedings 5, Springer, 2008, pp. 1--19.

\bibitem{hilton2002differential}
M.~Hilton, Differential privacy: a historical survey, Cal Poly State University (2002).

\bibitem{renyi1961measures}
A.~R{\'e}nyi, On measures of entropy and information, in: Proceedings of the Fourth Berkeley Symposium on Mathematical Statistics and Probability, Volume 1: Contributions to the Theory of Statistics, Vol.~4, University of California Press, 1961, pp. 547--562.

\bibitem{fang2022improved}
H.~Fang, X.~Li, C.~Fan, P.~Li, Improved convergence of differential private sgd with gradient clipping, in: The Eleventh International Conference on Learning Representations, 2022.

\bibitem{lecun1998gradient}
Y.~LeCun, L.~Bottou, Y.~Bengio, P.~Haffner, Gradient-based learning applied to document recognition, Proceedings of the IEEE 86~(11) (1998) 2278--2324.

\bibitem{krizhevsky2009learning}
A.~Krizhevsky, G.~Hinton, et~al., Learning multiple layers of features from tiny images (2009).

\bibitem{brock2021high}
A.~Brock, S.~De, S.~L. Smith, K.~Simonyan, High-performance large-scale image recognition without normalization, in: International Conference on Machine Learning, PMLR, 2021, pp. 1059--1071.

\bibitem{zamiela2023thermal}
C.~Zamiela, W.~Tian, S.~Guo, L.~Bian, Thermal-porosity characterization data of additively manufactured ti--6al--4v thin-walled structure via laser engineered net shaping, Data in Brief 51 (2023) 109722.

\bibitem{khanzadeh2019situ}
M.~Khanzadeh, S.~Chowdhury, M.~A. Tschopp, H.~R. Doude, M.~Marufuzzaman, L.~Bian, In-situ monitoring of melt pool images for porosity prediction in directed energy deposition processes, IISE Transactions 51~(5) (2019) 437--455.

\bibitem{esfahani2022situ}
M.~N. Esfahani, M.~M. Bappy, L.~Bian, W.~Tian, In-situ layer-wise certification for direct laser deposition processes based on thermal image series analysis, Journal of Manufacturing Processes 75 (2022) 895--902.

\bibitem{tian2020physics}
Q.~Tian, S.~Guo, Y.~Guo, et~al., A physics-driven deep learning model for process-porosity causal relationship and porosity prediction with interpretability in laser metal deposition, CIRP Annals 69~(1) (2020) 205--208.

\bibitem{bappy2022situ}
M.~M. Bappy, C.~Liu, L.~Bian, W.~Tian, In-situ layer-wise certification for direct energy deposition processes based on morphological dynamics analysis, Journal of Manufacturing Science and Engineering (2022) 1--35.

\bibitem{ye2021situ}
Z.~Ye, C.~Liu, W.~Tian, C.~Kan, In-situ point cloud fusion for layer-wise monitoring of additive manufacturing, Journal of Manufacturing Systems 61 (2021) 210--222.

\bibitem{seifi2019layer}
S.~H. Seifi, W.~Tian, H.~Doude, M.~A. Tschopp, L.~Bian, Layer-wise modeling and anomaly detection for laser-based additive manufacturing, Journal of Manufacturing Science and Engineering 141~(8) (2019) 081013.

\bibitem{al2020effects}
A.~Y. Al-Maharma, S.~P. Patil, B.~Markert, Effects of porosity on the mechanical properties of additively manufactured components: a critical review, Materials Research Express 7~(12) (2020) 122001.

\bibitem{sola2019microstructural}
A.~Sola, A.~Nouri, Microstructural porosity in additive manufacturing: The formation and detection of pores in metal parts fabricated by powder bed fusion, Journal of Advanced Manufacturing and Processing 1~(3) (2019) e10021.

\bibitem{sanaei2021defects}
N.~Sanaei, A.~Fatemi, Defects in additive manufactured metals and their effect on fatigue performance: A state-of-the-art review, Progress in Materials Science 117 (2021) 100724.

\bibitem{ma2018shufflenet}
N.~Ma, X.~Zhang, H.-T. Zheng, J.~Sun, Shufflenet v2: Practical guidelines for efficient cnn architecture design, in: Proceedings of the European conference on computer vision (ECCV), 2018, pp. 116--131.

\bibitem{Torch}
\url{https://pytorch.org}, [accessed on 12-Jan-2023] (2017).

\bibitem{yousefpour2021opacus}
A.~Yousefpour, I.~Shilov, A.~Sablayrolles, D.~Testuggine, K.~Prasad, M.~Malek, J.~Nguyen, S.~Ghosh, A.~Bharadwaj, J.~Zhao, et~al., Opacus: User-friendly differential privacy library in pytorch, arXiv preprint arXiv:2109.12298 (2021).

\bibitem{bagdasaryan2019differential}
E.~Bagdasaryan, O.~Poursaeed, V.~Shmatikov, Differential privacy has disparate impact on model accuracy, Advances in neural information processing systems 32 (2019).

\bibitem{hassan2019differential}
M.~U. Hassan, M.~H. Rehmani, J.~Chen, Differential privacy techniques for cyber physical systems: a survey, IEEE Communications Surveys \& Tutorials 22~(1) (2019) 746--789.

\bibitem{jiang2021differential}
B.~Jiang, J.~Li, G.~Yue, H.~Song, Differential privacy for industrial internet of things: Opportunities, applications, and challenges, IEEE Internet of Things Journal 8~(13) (2021) 10430--10451.

\bibitem{gartner2022local}
S.~G{\"a}rtner, M.~Oberle, Local differential privacy in smart manufacturing: Application scenario, mechanisms and tools, in: Proceedings of the Conference on Production Systems and Logistics: CPSL 2022, Hannover: publish-Ing., 2022, pp. 482--491.

\bibitem{jain2018differential}
P.~Jain, M.~Gyanchandani, N.~Khare, Differential privacy: its technological prescriptive using big data, Journal of Big Data 5~(1) (2018) 1--24.

\bibitem{balletti20173d}
C.~Balletti, M.~Ballarin, F.~Guerra, 3d printing: State of the art and future perspectives, Journal of Cultural Heritage 26 (2017) 172--182.

\bibitem{jandyal20223d}
A.~Jandyal, I.~Chaturvedi, I.~Wazir, A.~Raina, M.~I.~U. Haq, 3d printing--a review of processes, materials and applications in industry 4.0, Sustainable Operations and Computers 3 (2022) 33--42.

\bibitem{fullington2023design}
D.~Fullington, L.~Bian, W.~Tian, Design de-identification of thermal history for collaborative process-defect modeling of directed energy deposition processes, Journal of Manufacturing Science and Engineering 145~(5) (2023) 051004.

\bibitem{owusu2023msdp}
K.~Owusu-Agyemeng, Z.~Qin, H.~Xiong, Y.~Liu, T.~Zhuang, Z.~Qin, Msdp: multi-scheme privacy-preserving deep learning via differential privacy, Personal and Ubiquitous Computing (2023) 1--13.

\bibitem{dwork2014algorithmic}
C.~Dwork, A.~Roth, et~al., The algorithmic foundations of differential privacy, Foundations and Trends{\textregistered} in Theoretical Computer Science 9~(3--4) (2014) 211--407.

\bibitem{narayanan2019robust}
A.~Narayanan, V.~Shmatikov, Robust de-anonymization of large sparse datasets: a decade later, May 21 (2019) 2019.

\bibitem{gil2019understanding}
E.~Gil~Gonz{\'a}lez, P.~De~Hert, Understanding the legal provisions that allow processing and profiling of personal data—an analysis of gdpr provisions and principles, in: Era Forum, Vol.~19, Springer, 2019, pp. 597--621.

\end{thebibliography}

\end{document}